\documentclass[lettersize,journal]{IEEEtran}
\usepackage{amsmath,amsfonts}
\usepackage{algorithmic}
\usepackage{algorithm}
\usepackage{array}
\usepackage[caption=false,font=normalsize,labelfont=sf,textfont=sf]{subfig}
\usepackage{textcomp}
\usepackage{stfloats}
\usepackage{url}
\usepackage{verbatim}
\usepackage{graphicx}
\usepackage{cite}
\usepackage{booktabs}
\usepackage{amssymb} 
\usepackage{adjustbox}
\usepackage{enumitem}

\usepackage{xcolor}
\usepackage{subcaption}
\usepackage{adjustbox}
\usepackage{amsfonts}

\usepackage{mathtools}
\PassOptionsToPackage{table}{xcolor}  
\usepackage{colortbl}
\begin{document}

\title{GLASS: Geometry-aware Local Alignment and Structure Synchronization Network for 2D-3D Registration}

\author{
  Zhixin Cheng, Jiacheng Deng, Xinjun Li, Bohao Liao, Li Liu, Xiaotian Yin, Baoqun Yin, Tianzhu Zhang\textsuperscript{*}%
  \thanks{\textsuperscript{*}Corresponding author: tzzhang@ustc.edu.cn.}%
  \thanks{Zhixin Cheng, Xinjun Li, Bohao Liao, Baoqun Yin and Tianzhu Zhang are with the School of Information Science and Technology, University of Science and Technology of China, Hefei 230027, China (e-mail: chengzhixin@mail.ustc.edu.cn; lxj3017@mail.ustc.edu.cn; liaobh@mail.ustc.edu.cn; bqyin@ustc.edu.cn; tzzhang@ustc.edu.cn).}%
  \thanks{Li Liu and Xiaotian Yin are with the Institute of Advanced Technology, University of Science and Technology of China, Hefei 230027, China (e-mail: liu\_li@mail.ustc.edu.cn; xiaotianyin@mail.ustc.edu.cn).}%
  \thanks{Jiacheng Deng is with Meituan Inc, Beijing 100000, China (e-mail: dengjc@mail.ustc.edu.cn).}
  
  \thanks{Manuscript received April 19, 2021; revised August 16, 2021.}
}

\markboth{Journal of \LaTeX\ Class Files,~Vol.~14, No.~8, August~2021}%
{Shell \MakeLowercase{\textit{et al.}}: A Sample Article Using IEEEtran.cls for IEEE Journals}


\maketitle

\begin{abstract}
Image-to-point cloud registration methods typically follow a coarse-to-fine pipeline, extracting patch-level correspondences and refining them into dense pixel-to-point matches. However, in scenes with repetitive patterns, images often lack sufficient 3D structural cues and alignment with point clouds, leading to incorrect matches. Moreover, prior methods usually overlook structural consistency, limiting the full exploitation of correspondences. To address these issues, we propose two novel modules: the Local Geometry Enhancement (LGE) module and the Graph Distribution Consistency (GDC) module. LGE enhances both image and point cloud features with normal vectors, injecting geometric structure into image features to reduce mismatches. GDC constructs a graph from matched points to update features and explicitly constrain similarity distributions. Extensive experiments and ablations on two benchmarks, RGB-D Scenes v2 and 7-Scenes, demonstrate that our approach achieves state-of-the-art performance in image-to-point cloud registration.
\end{abstract}


\section{Introduction}
Image-to-point cloud registration aims to estimate a rigid transformation that aligns a 3D point cloud with a 2D image captured from the same scene. This involves establishing cross-modal correspondences between pixels and 3D points, followed by pose estimation to recover rotation and translation. The task plays a critical role in various applications such as 3D reconstruction \cite{3dreconstruction,5,6,deng2}, SLAM \cite{slam,7,8,deng}, and visual localization \cite{visuallocalization,3,diffreg}. However, the inherently different structures of images and point clouds pose a major challenge: images are dense and structured as 2D grids, while point clouds are sparse, unordered, and irregular in 3D space. This large modality gap complicates the extraction of meaningful cross-modal correspondences, hindering effective alignment between the two data types.

\begin{figure}[!t]
\centering
\includegraphics[width=\columnwidth]{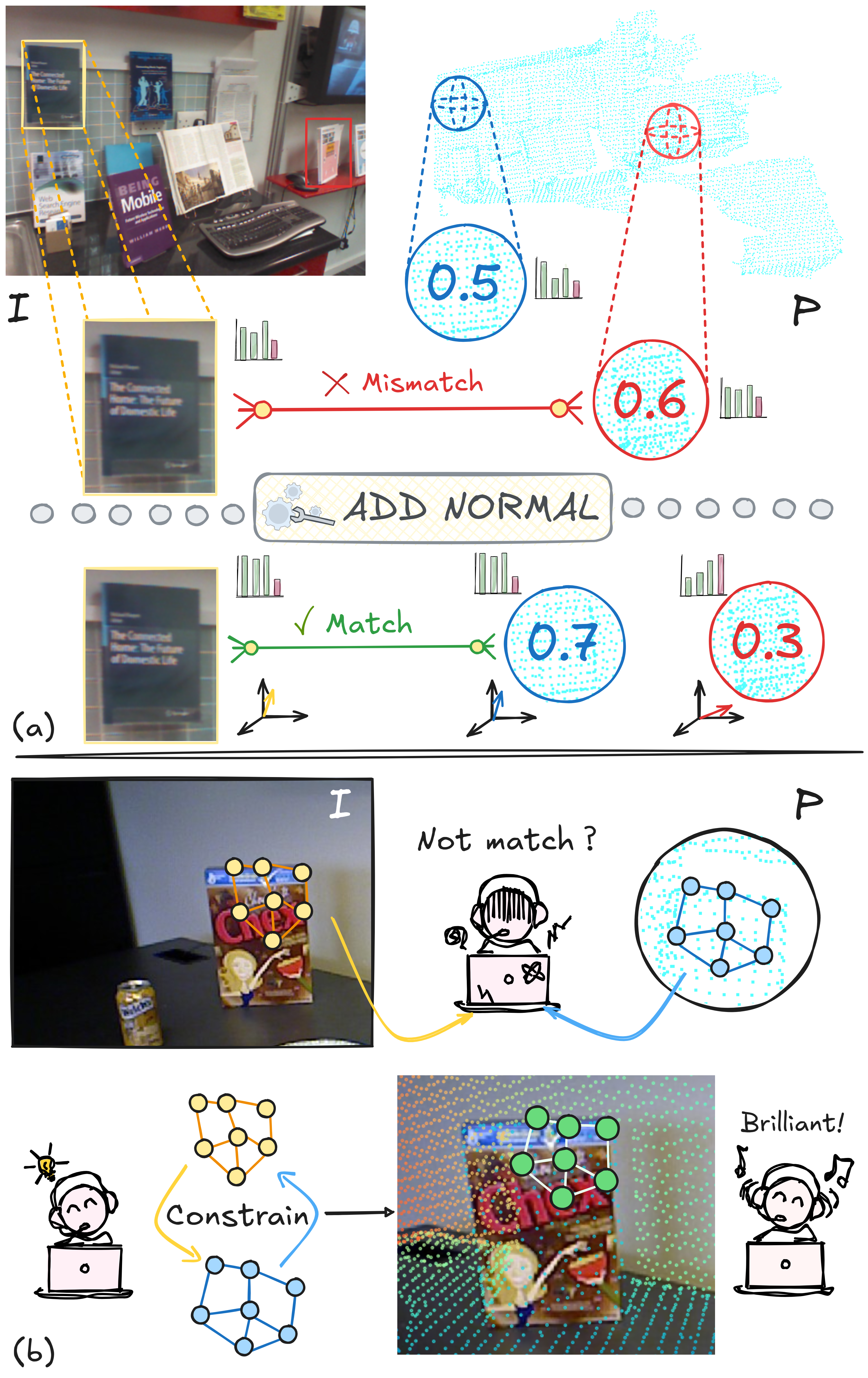} 
\vspace{-15pt}
\caption{(a) Visualization of reduced mismatches aided by normal-based structural information. The normals provide shared attributes between the image and the point cloud, enhancing the matching accuracy between them.
(b) Visualization of structural similarity distribution constraint. By enforcing the structural consistency constraint, the alignment of image and point cloud keypoints is improved, leading to a better understanding of the scene.
}
\label{fig1}
\end{figure}

To tackle the image-to-point cloud registration problem, existing approaches generally fall into two categories. The first is the detect-then-match approach \cite{2d3dmatchnet, p2, corri2p,9,10}, which detects 2D and 3D key points in the image and point cloud separately and then matches them based on semantic features. This approach often struggles due to the modality gap: image keypoints depend on texture, while point cloud keypoints rely on geometry, making repeatable detection difficult. Moreover, the large discrepancy between 2D and 3D feature representations makes it hard to learn consistent descriptors. To mitigate these issues, recent works explore detection-free pipelines \cite{cofii2p,4,1,2}. A representative method, 2D3D-MATR \cite{matr2d3d}, introduces a coarse-to-fine framework that first performs patch-level matching and then refines correspondences to the point level, followed by rigid transformation estimation via PnP-RANSAC \cite{pnp,ransac}. However, under challenging conditions such as illumination changes, low-texture regions, or repetitive patterns, the lack of structural guidance can still lead to inaccurate cross-modal matching.

Based on the discussion above, we identify two key challenges that must be addressed for accurate and reliable image-to-point cloud matching. \textbf{Firstly, how to incorporate additional 3D information at the feature level to enhance structural representation and reduce cross-modal mismatches.} Depth maps are commonly used to provide 3D cues, but their scale ambiguity limits their effectiveness for local matching. In contrast, surface normals are translation- and scale-invariant, making them more suitable for matching. We leverage normals to enhance the 3D geometric awareness of image features, improving their alignment with point cloud features. As shown in Fig.~\ref{fig1}(a), original 2D features may cause incorrect matches due to weak visual cues, while the structural guidance from normals helps suppress such errors and improves matching reliability.
\textbf{Secondly, how to build structured similarity constraints at the matching level to fully exploit correspondences and enforce cross-modal consistency.} While coarse-to-fine pipelines achieve accurate pixel-to-point correspondences, they often overlook the structural relations among neighboring matches. Constructing a graph from matched points allows aggregation of local geometry and helps reduce domain gaps. As illustrated in Fig.~\ref{fig1}(b), once pixel-point matches are established, their graph structures should be aligned. The graph highlights matching errors and, when constrained, promotes structural consistency, thereby improving overall registration accuracy.

To tackle the above challenges, we propose Geometry-aware Local Alignment and Structure Synchronization Network for 2D-3D Registration (GLASS), a novel framework for 2D-3D registration. It consists of two key modules: Local Geometry Enhancement (LGE) module and Graph Distribution Consistency (GDC) module.
In the LGE, we integrate surface normals into image features to provide 3D structural context and improve their correspondence with point cloud features. Surface normals for images are derived from predicted depth maps, while point cloud normals are directly computed from 3D coordinates. By fusing raw 2D and 3D descriptors with normal-based features, we significantly improve the cross-modal discriminative power under challenging conditions. Notably, the pseudo-normal labels for training are generated from depth maps predicted by Depth Anything v2 \cite{deptha}, requiring no additional manual annotation.
In the GDC, we construct a graph between matched pixels and points within local neighborhoods and apply a graph neural network \cite{gnn,gat} to aggregate information and constrain the structured similarity distribution. This graph structure captures both local and global relationships, helping reduce domain gaps and improving robustness. By maximizing the structural alignment between accurately matched cross-modal graphs, GDC effectively suppresses many-to-one mismatches and contributes to the precise registration of images and point clouds.

In summary, our work can be outlined as follows:

\begin{itemize}
    \item We design a novel Geometry-aware Local Alignment and Structure Synchronization Network that achieves high accuracy and generalization for 2D-3D registration.
    \item We propose a Local Geometry Enhancement module to inject 3D awareness via surface normals, strengthening the structural correlation between image and point cloud features. We also introduce a Graph Distribution Consistency module that builds graph structures from correspondences and enforces structural similarity constraints to improve cross-modal consistency.
    \item Validated by comprehensive experiments and ablations, our method achieves state-of-the-art performance on RGB-D Scenes v2 and 7-Scenes.
\end{itemize}

\section{Related Works}

In this section, we briefly overview related works on I2P registration, including stereo image registration, point cloud registration, and inter-modality registration. 

\textbf{Image Registration.}
Conventional stereo image registration primarily depends on detector-based approaches that leverage handcrafted keypoint detection and description to establish feature correspondences. Classical methods such as SIFT \cite{sift} and ORB \cite{orb} have been widely adopted for 2D image matching tasks. With the advent of deep learning, learning-based methods like SuperGlue \cite{superglue} have significantly enhanced matching accuracy by integrating transformer architectures \cite{transformer}. Despite their advances, these methods often struggle in regions lacking salient features, resulting in limited robustness due to low keypoint repeatability. To overcome this limitation, detector-free approaches such as LoFTR \cite{loftr} and Efficient LoFTR \cite{efficientloftr} have been proposed. These methods adopt a coarse-to-fine matching strategy with global receptive fields, enabling more reliable and dense correspondence estimation without relying on explicit keypoint detection.

\textbf{Point Cloud Registration.}
Point cloud registration has progressed from early handcrafted descriptors such as PPF \cite{ppf} and FPFH \cite{fpfh} to modern deep learning-based techniques. A notable milestone is CoFiNet \cite{cofinet}, which introduced a coarse-to-fine pipeline and set the foundation for detector-free registration methods. More recent efforts have aimed to replace classical estimators like RANSAC \cite{ransac} with learning-based alternatives that offer improved speed and accuracy. GeoTransformer \cite{geotransformer} further advances this direction by leveraging transformers to model global structural dependencies, significantly boosting inlier ratios. Additionally, it proposes a local-to-global registration (LGR) framework that achieves accurate alignment without relying on RANSAC.

\textbf{Inter-Modality Registration.}
Compared to intra-modality registration, inter-modality registration poses greater challenges due to larger differences in data domains. Traditional methods typically follow a detect-then-match paradigm. For example, 2D3DMatch-Net \cite{2d3dmatchnet} uses SIFT \cite{sift} for keypoint extraction, while ISS \cite{iss} constructs local patches on the point cloud and employs CNNs and PointNet \cite{pointnet} for descriptor extraction and matching. P2-Net \cite{p2} further improves efficiency by jointly detecting and matching keypoints in a single step. However, in cross-modal scenarios, the accuracy of keypoint detection often degrades \cite{keypoint,keypoint2}, limiting overall performance and leading to the development of detector-free methods. 2D3D-MATR \cite{matr2d3d} adopts a coarse-to-fine strategy: it first performs coarse matching using a transformer network, followed by refinement via PnP-RANSAC, avoiding keypoint detection and improving descriptor alignment. FreeReg \cite{freereg} utilizes features from pretrained diffusion models and geometric cues from depth maps to achieve accurate image-to-point cloud registration without task-specific training. B2-3D \cite{cheng11} improves registration accuracy through uncertainty modeling and domain adaptation.
CA-I2P \cite{cai2p} introduces channel adaptation and global optimal selection to better align cross-modal features and reduce redundant matches, achieving improved registration accuracy.
Flow-I2P \cite{flowi2p} improves image-to-point-cloud registration by using Beltrami flow for better manifold alignment and enhanced registration accuracy.
Based on 2D3D-MATR, our proposed GLASS method also adopts a detector-free design and incorporates normal vector information and graph-based constraints to alleviate mismatches and structural bias, achieving a new state-of-the-art in image-to-point cloud registration. 

While surface normal estimation~\cite{01,02,03} and graph-based regularization~\cite{04,05,06} have been widely studied in classical vision tasks, most existing approaches apply these techniques within the same modality (e.g., image-to-image or point-to-point registration), where data distribution and spatial structure are relatively consistent. In contrast, cross-modal registration tasks (e.g., from RGB images to 3D point clouds) introduce distinct challenges, including geometric distortions and heterogeneous feature representations. To the best of our knowledge, there remains a lack of systematic research that explicitly addresses the modality gap between 2D pixel space and 3D geometric space. In this work, we focus on organically integrating normal estimation with graph regularization, and explore how these classical components can be adapted for cross-modal registration scenarios. This integration-oriented approach, which adapts well-established techniques to mitigate modality gaps, serves as a modest but meaningful complement to prior studies that primarily concentrate on single-modality settings.

\begin{figure*}[t]
\centering
\includegraphics[width=\textwidth]{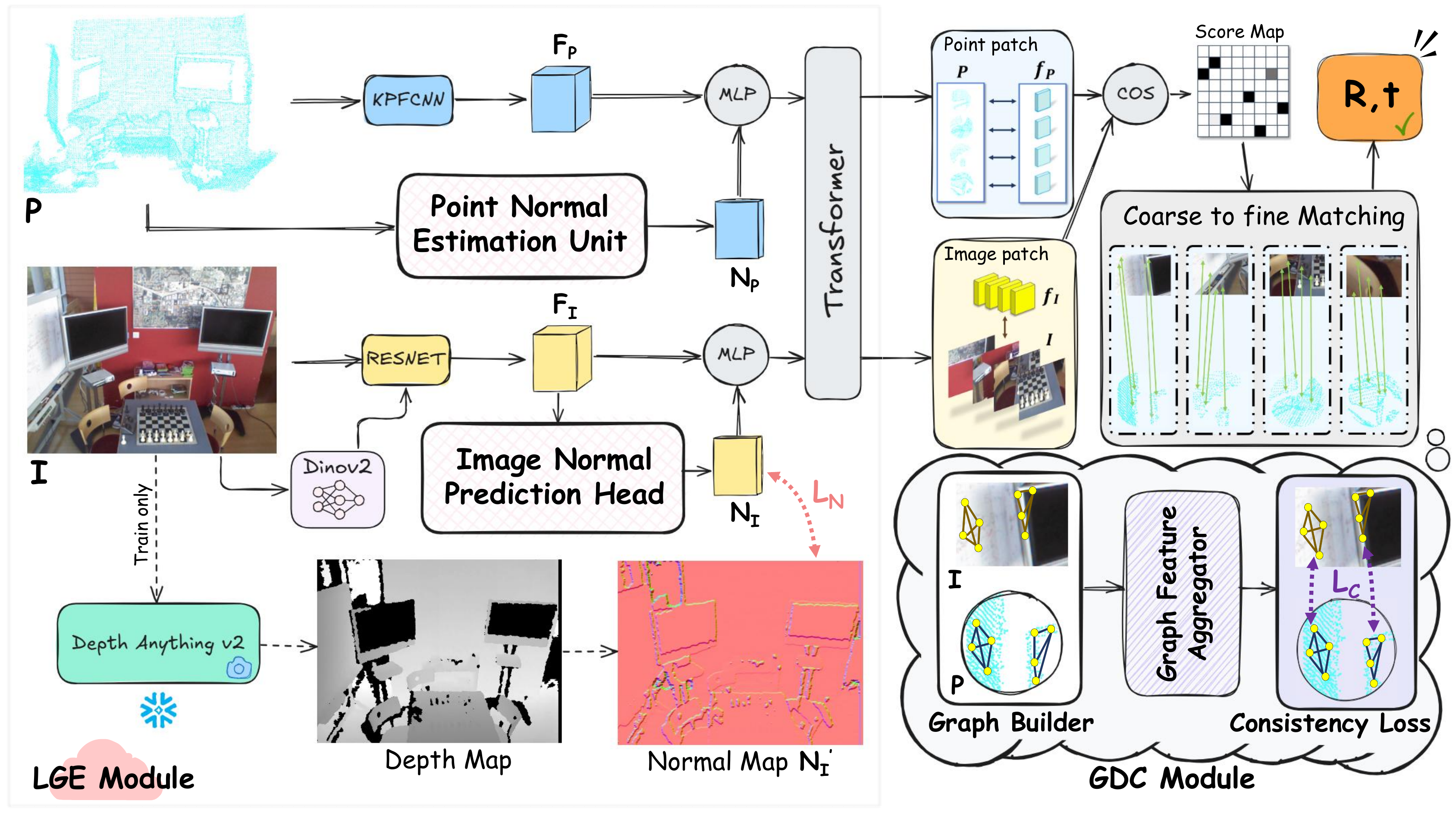} 
\caption{Overall pipeline of the GLASS. It includes the Local Geometry Enhancement (LGE) and Graph Distribution Consistency (GDC) modules. In LGE, the image branch predicts surface normals via an Image Normal Prediction Head, trained using pseudo normal labels generated by Depth Anything v2. The point cloud branch computes normals using a Point Normal Estimation Unit. By fusing normals with initial features, the model gains structural awareness and reduces mismatches. In GDC, matching points in the local neighborhoods of both image and point cloud are constructed as graphs to update features. Similarity distributions are then constrained to enforce structural consistency, improving the quality of dense correspondences.}
\label{fig2}
\end{figure*}

\section{Method}
\subsection{Overall}
Let $I \in \mathbb{R}^{H \times W \times 3}$ and $P \in \mathbb{R}^{N \times 3}$ represent an image and a point cloud from the same scene, where $H$ and $W$ denote the height and width of the image, and $N$ is the number of 3D points. The goal of image-to-point cloud registration is to estimate a rigid transformation $[R, \mathbf{t}]$, where the rotation matrix $R \in \mathrm{SO}(3)$ and the translation vector $\mathbf{t} \in \mathbb{R}^3$.

Our method, GLASS (shown in Fig.~\ref{fig2}), adopts a detector-free registration paradigm. The Local Geometry Enhancement (LGE) module introduces surface normal information to provide additional 3D cues from images, enhancing the feature correlation between the image and point cloud while effectively reducing matching errors. After obtaining coarse patch-level correspondences, the Graph Distribution Consistency (GDC) module imposes structural constraints to enhance the quality of dense matches. Finally, the PnP+RANSAC algorithm is used to robustly estimate the rigid transformation.

\subsection{Local Geometry Enhancement Module}
We use ResNet \cite{resnet} with an FPN \cite{fpn} to extract features from images and KPFCNN \cite{kpfcnn} to extract features from point clouds. 
We augment the 2D and 3D features with their positional information before the attention layer
\begin{equation}
f^{\mathcal{I}}_{\text{pos}} = f_i + \phi(I), \qquad f^{\mathcal{P}}_{\text{pos}} = f_p + \phi(P).
\end{equation}

The Fourier embedding function \(\phi(x)\) \cite{fourier} encodes positional information by transforming it into a sequence of sine and cosine terms:
\begin{multline}
\phi(x) = \big[x, \sin(2^0 x), \cos(2^0 x), \ldots, \\
\sin(2^{L-1} x), \cos(2^{L-1} x)\big],
\end{multline}
where \(L\) is the length of the embedding. This transformation incorporates spatial positioning into the features.
After obtaining the positional encodings, the first two spatial dimensions of the 2D features are flattened, and the positional encodings \( f^{\mathcal{I}}_{\text{pos}} \) and \( f^{\mathcal{P}}_{\text{pos}} \) are then added to the features for subsequent processing. For the image branch, representations from DINO v2 \cite{dinov2} are further integrated to enrich the features. At the lowest resolution, the image features are denoted as $\mathbf{F}_I \in \mathbb{R}^{h \times w \times c}$, while the point cloud features are represented as $\mathbf{F}_P \in \mathbb{R}^{n \times c}$. We hope to use surface normals as structure-aware cues to bridge image and point cloud representations, thereby minimizing cross-modal matching errors.

Thanks to the inherent 3D geometric structure of point clouds, the surface normal feature $\mathbf{N}_P$ can be directly estimated from neighboring point coordinates within the Point Normal Estimation Unit. For each point $\mathbf{p}_a \in \mathbb{R}^3$ in the point cloud, the surface normal $\mathbf{N}_P(a)$ is estimated based on its $k$ nearest neighbors $\mathcal{N}_a$. First, the neighborhood centroid is computed as 
\begin{equation}
\bar{\mathbf{p}}_a = \frac{1}{k} \sum_{\mathbf{p}_b \in \mathcal{N}_a} \mathbf{p}_b,
\vspace{-2pt}
\end{equation}
and the covariance matrix of the neighborhood is constructed as
\begin{equation}
\mathbf{C}_a = \frac{1}{k} \sum_{\mathbf{p}_b \in \mathcal{N}_a} (\mathbf{p}_b - \bar{\mathbf{p}}_a)(\mathbf{p}_b - \bar{\mathbf{p}}_a)^\top,
\vspace{-2pt}
\end{equation}
The normal vector is then obtained as the eigenvector corresponding to the smallest eigenvalue of $\mathbf{C}_a$, i.e.,
\begin{equation}
\mathbf{C}_a \mathbf{N}_P(a) = \lambda_{\min} \mathbf{N}_P(a),
\end{equation}
and is finally normalized as 
\begin{equation}
\mathbf{N}_P(a) \leftarrow \mathbf{N}_P(a) / |\mathbf{N}_P(a)|.
\end{equation}

Unlike point clouds, estimating surface normals from images requires additional processing. Surface normals characterize the local orientation of surfaces and serve as 3D signals with inherent translation and scale invariance. During training, we generate supervision for image normals using the monocular depth estimation model Depth Anything v2~\cite{deptha}. Although monocular depth inherently suffers from scale ambiguity, converting depth maps into surface normals mitigates this issue by providing robust geometric cues that are invariant to scale and shift, thereby improving feature matching.

Given an input image $I$, a predicted depth map $D_I$ is used to estimate surface normals. For a pixel $A(u, v)$, where $(u, v) \in [0, h] \times [0, w]$, the depth gradients are approximated using finite differences:
\begin{equation}
\frac{\partial D_I}{\partial u} \approx D_I(u+1,v) - D_I(u-1,v),
\end{equation}
\begin{equation}
\frac{\partial D_I}{\partial v} \approx D_I(u,v+1) - D_I(u,v-1).
\end{equation}
Assuming the 3D coordinate of the pixel is $(u, v, D_I(u,v))$, the corresponding image normal $\mathbf{N}_I'$ is computed as:
\begin{equation}
\mathbf{N}_I' =
\frac{
\left(
-\frac{\partial D_I}{\partial u},\
-\frac{\partial D_I}{\partial v},\
1
\right)
}{
\left|
\left(
-\frac{\partial D_I}{\partial u},\
-\frac{\partial D_I}{\partial v},\
1
\right)
\right|
},
\end{equation}
where the denominator normalizes the vector to unit length. The resulting $\mathbf{N}_I'$ encodes the local surface orientation at $A(u,v)$, capturing the underlying 3D geometry of the scene.

To reduce computational overhead, we introduce an Image Normal Prediction Head, implemented as a shallow MLP, which directly predicts surface normals $\mathbf{N}_I$ from image features $\mathbf{F}_I$. During training, the predicted normals are supervised by the estiamtion normals $\mathbf{N}_I'$ derived from Depth Anything v2. At inference time, the depth estimation network is disabled to avoid additional computational cost.
The normal supervision loss is defined as:
\vspace{-1pt}
\begin{equation}
\mathcal{L}_{N} = 1 - \frac{1}{hw} \sum_{d=1}^{hw} \left( \mathbf{N}_I \cdot \mathbf{N}_I' \right),
\end{equation}
where $\mathbf{N}_I$ and $\mathbf{N}_I'$ denote the predicted normals and the pseudo ground-truth normal labels at pixel $d$, respectively, and $hw$ is the total number of valid pixel positions. This loss encourages the predicted normals to align with the underlying 3D geometry, thereby enhancing 2D–3D consistency while preserving inference efficiency.

For the image features $\mathbf{F}_I$ and normals $\mathbf{N}_I$, as well as the point cloud features $\mathbf{F}_P$ and normals $\mathbf{N}_P$, we employ lightweight MLPs to align their dimensions and fuse them before feeding the results into the transformer for cross-modal interaction. This facilitates adaptive and structure-aware alignment between image and point cloud representations. Since surface normals encode local geometric orientations and exhibit inherent cross-modal correlations, they provide additional 3D cues to image features, reducing mismatches and significantly improving the robustness and accuracy of image-to-point cloud registration.

\begin{algorithm}[ht]
\footnotesize
\caption{Normal computation and constraint for image and point cloud}
\label{alg:normals}

\textbf{Input:} Point cloud $\mathbf{P}$, Image $\mathbf{I}$

\textbf{Output:} Point cloud normals $\mathbf{N}_p$, Image normals $\mathbf{N}_I$, Loss $\mathcal{L}_N$

\textbf{Point cloud normals:}\\
For each point $p_a \in \mathbf{P}$:
\begin{itemize}
  \item $\mathcal{N}_a \leftarrow k$-NN($p_a$)
  \item $\mathbf{C}_a \leftarrow \tfrac{1}{k}\sum (p_b-\bar{p}_a)(p_b-\bar{p}_a)^\top$
  \item $\mathbf{N}_p(a) \leftarrow \text{normalize}(\text{eigenvector}_{\min}(\mathbf{C}_a))$
\end{itemize}

\textbf{Image normals:}\\
$D_i \leftarrow \text{DepthAnything}(\mathbf{I})$
For each pixel $(u,v)$:
\begin{itemize}
  \item $\partial D/\partial u, \partial D/\partial v \leftarrow$ finite diff.
  \item $\mathbf{N}_I'(u,v) \leftarrow \text{normalize}((- \partial D/\partial u,-\partial D/\partial v,1))$
  \item $\mathbf{N}_I(u,v) \leftarrow \text{MLP}(f_i(u,v))$
\end{itemize}

\textbf{Normal supervision:}\\
$\mathcal{L}_N \leftarrow 1-\tfrac{1}{hw}\sum (\mathbf{N}_I\cdot\mathbf{N}_I')$

\end{algorithm}

\subsection{Graph Distribution Consistency Module}

After transformer-based interaction, coarse-level matches between image and point cloud features are obtained via a cosine similarity score map, followed by fine-grained feature matching to establish point-level correspondences. These correspondences can be treated as keypoints with inherent matching relations, which, if further exploited by incorporating their local structural context, can help reduce errors and better leverage the correspondence consistency.

To enhance the structural awareness of image and point cloud features after coarse-level matching, we construct k-NN graphs over the matched keypoints in both modalities. Specifically, we construct a 2D graph by treating matched image keypoints as nodes and connecting each to its $k$ nearest neighbors based on pixel coordinates. Similarly, a 3D graph is built by linking each matched point cloud node to its $k$ nearest neighbors in 3D space. These graphs capture local geometric structure for subsequent feature aggregation. These graphs capture local neighborhood structures and allow us to model intra-modality spatial relationships, which further help reduce the domain gap between image and point cloud features by enforcing structural consistency within each modality.

\begin{algorithm}[ht]
\footnotesize
\caption{Cross-Modal Feature Refinement and LightGAT Aggregation}
\label{alg:feature_refinement}

\textbf{Input:} Image features $\mathbf{F}_I$, Point cloud features $\mathbf{F}_P$, Image keypoints $\mathbf{K}_I$, Point cloud keypoints $\mathbf{K}_P$, k-NN parameter $k$, warm-up epochs $E_{w}$, total epochs $E_{t}$

\textbf{Output:} Refined image features $\mathbf{F}_I^{ref}$, Refined point cloud features $\mathbf{F}_P^{ref}$, Loss $\mathcal{L_C}$

\textbf{Coarse-Level Matching:}\\
$\mathbf{S} \leftarrow \text{cosine\_similarity}(\mathbf{F}_I, \mathbf{F}_P)$

\textbf{k-NN Graph Construction:}\\
$\mathbf{G}_I \leftarrow \text{k-NN}(\mathbf{K}_I, k)$ \quad \text{(2D graph for image keypoints)}\\
$\mathbf{G}_P \leftarrow \text{k-NN}(\mathbf{K}_P, k)$ \quad \text{(3D graph for point cloud keypoints)}

\textbf{LightGAT Feature Aggregation:}\\
\textit{For each node $n_i$ in image graph $\mathbf{G}_I$}:\\
\quad $\mathbf{A}_I \leftarrow \text{compute\_attention}(n_i, \text{neighbors of } n_i)$\\
\quad $\mathbf{F}_I^{agg}(n_i) \leftarrow \text{update\_node\_features}(\mathbf{A}_I, \text{neighbors of } n_i)$

\textit{For each node $n_j$ in point cloud graph $\mathbf{G}_P$}:\\
\quad $\mathbf{A}_P \leftarrow \text{compute\_attention}(n_j, \text{neighbors of } n_j)$\\
\quad $\mathbf{F}_P^{agg}(n_j) \leftarrow \text{update\_node\_features}(\mathbf{A}_P, \text{neighbors of } n_j)$

\textbf{Fusion of Original and Refined Features:}\\
\textit{For each pixel $(u, v)$ in the image:}\\
\quad $\mathbf{F}_I^{ref}(u,v) \leftarrow \text{MLP}(\mathbf{F}_I(u,v), \mathbf{F}_I^{agg}(u,v))$

\textit{For each point $p_k$ in the point cloud:}\\
\quad $\mathbf{F}_P^{ref}(p_k) \leftarrow \text{MLP}(\mathbf{F}_P(p_k), \mathbf{F}_P^{agg}(p_k))$

\textbf{Similarity Distribution and Loss Calculation:}\\
$\mathbf{S}_I \leftarrow \text{compute\_similarity\_matrix}(\mathbf{F}_I^{ref})$\\
$\mathbf{S}_P \leftarrow \text{compute\_similarity\_matrix}(\mathbf{F}_P^{ref})$\\
$\mathcal{L_C} \leftarrow \text{Frobenius\_norm}(\mathbf{S}_I - \mathbf{S}_P)$

\textbf{Training Loop with Warm-Up:}\\
\textit{For epoch = 1 to $E_{t}$:}\\
\quad \textbf{If} $epoch \leq E_{w}$: \\
\quad \quad $\mathcal{L_C} \leftarrow \text{compute\_loss}(\mathbf{F}_I, \mathbf{F}_P)$ \quad \text{(focusing on learning correspondences)}\\
\quad \textbf{Else:}\\
\quad \quad $\mathbf{F}_I^{ref}, \mathbf{F}_P^{ref} \leftarrow \text{refine\_features}(\mathbf{G}_I, \mathbf{G}_P, \mathbf{F}_I, \mathbf{F}_P)$\\
\quad \quad $\mathcal{L_C} \leftarrow \text{compute\_loss}(\mathbf{F}_I^{ref}, \mathbf{F}_P^{ref})$ \quad \text{(apply structural constraints)}\\
\quad $\text{optimize}(\mathcal{L_C})$

\end{algorithm}

We then apply a lightweight Graph Attention Network (LightGAT) \cite{lightgat,lightgat1} to aggregate features within each constructed graph. The graph is defined such that each node corresponds to a feature vector from either the image or the point cloud, and the edges capture the relationships between nodes based on spatial proximity, semantic similarity, or other contextual factors.

For each node in the graph, LightGAT computes attention weights over its neighbors using a self-attention mechanism. This process begins by projecting both the node’s feature and the feature of each neighboring node into a shared attention space via a learnable linear transformation. The attention scores are then computed using a compatibility function, typically a scaled dot-product between the transformed node features. These attention scores are normalized through a softmax operation, ensuring that the sum of attention weights for each node’s neighbors equals one, which allows the model to focus on the most relevant neighboring nodes.
The node features are updated through a weighted sum of neighboring features, where the weights are the computed attention scores. This operation allows each node to selectively aggregate information from its neighbors, enabling the network to emphasize the most relevant structural information while filtering out less informative neighbors. The update of node features thus incorporates a structural-aware refinement, meaning that the features are updated based on both the local geometry and the attention-based relationships within the graph.

To preserve the original information while incorporating contextual cues from the graph structure, the refined node features are adaptively fused with the original features. This fusion is achieved through a gated Multi-Layer Perceptron (MLP)-based fusion module, which computes a dynamic weighting for each feature. The gated fusion mechanism enables the model to learn when to rely more heavily on the refined features or the original ones, depending on the context, which promotes a better balance between preserving the original feature representation and enhancing it with structural context.
The resulting feature representations are more robust and geometry-consistent, making them well-suited for subsequent fine-level matching tasks. To ensure that the image and point cloud features align well, we compute the similarity distribution for the refined features and impose a distribution alignment constraint across modalities. This constraint enforces consistency in the feature distributions between the image and point cloud domains, thereby enhancing cross-modal compatibility and ensuring that the features are comparable in terms of their geometry.

 
We introduce a loss based on their pairwise similarity matrices. Given the normalized image and point cloud features $f_i \in \mathbb{R}^{m \times c}$ and $f_p \in \mathbb{R}^{m \times c}$, where $m$ denotes the number of matched pairs, we compute their self-similarity matrices as:
\begin{equation}
\mathbf{S}_I = f_i f_i^\top, \quad 
\mathbf{S}_P = f_p f_p^\top,
\end{equation}
the loss is defined as the Frobenius norm \cite{frobenorm} of the difference between these two matrices:
\begin{equation}
\mathcal{L}_C = \left\| \mathbf{S}_I - \mathbf{S}_P \right\|_F^2.
\end{equation}

Since accurate correspondences are not available at the early training stage, directly applying the distribution constraint may lead to instability. To address this, we adopt a warm-up strategy that initially allows the model to focus on learning reliable correspondences, and gradually introduces structural constraints to ensure training stability. This encourages the image and point cloud features to share similar internal relational structures, thereby improving cross-modal alignment.

The core motivation behind the GDC module stems from a fundamental challenge in cross-modal image-to-point cloud matching: reliable correspondence establishment depends not only on the similarity of individual feature pairs, but also on the global distributional structure of similarity scores across all candidates. Raw features from different modalities differ significantly due to sensing characteristics, appearance-geometry gaps, and encoder-specific biases, leading to a non-trivial cross-modal distribution shift. However, the relative relational structures (e.g., similarity ranking and local neighborhood topology) are comparatively invariant across modalities, as they are tightly coupled with the underlying scene geometry and semantics—even if absolute similarity magnitudes are misaligned. In cross-modal scenarios, the inherent modality gap (2D grid vs. 3D unordered points), scale inconsistencies, and geometric noise inevitably skew these global distributions, causing valid correspondences to lack sufficient separability from incorrect matches. This mismatch directly results in unstable confidence calibration and suboptimal correspondence selection, which is a root cause of failures in challenging regions (e.g., repetitive structures, low-texture areas). Against this backdrop, aligning similarity distributions is not just an empirical strategy, but a necessary step to address this fundamental issue.

Built upon the principle of relational invariance, the GDC module regularizes the similarity distribution (rather than directly forcing raw feature alignment) to reduce modality-induced bias in confidence estimation. It achieves this through two key principled mechanisms: (1) enforcing consistent similarity distribution characteristics across image and point cloud modalities to mitigate scoring bias; and (2) sharpening the contrast between positive and negative correspondences in a globally coherent manner to enhance the discriminability of valid matches. In other words, GDC moves beyond optimizing individual feature similarities to regularize the entire global similarity landscape of cross-modal features. This regularization not only makes the matching process more structurally consistent, but also stabilizes gradient signals during training, significantly improving the convergence robustness of the registration pipeline—especially in ambiguous regions where local cues are insufficient. Importantly, we acknowledge that relational structures are not always perfectly stable: in cases of severe sensing failures (e.g., strong depth artifacts, extremely sparse point clouds, or early training stages), the structure may become distorted. Additional experiments and analysis validating the robustness of our method under such challenging conditions are provided in the ablation studies.

\subsection{Loss Function}

Let us examine the loss functions for the coarse and fine-matching networks, which follow a coarse-to-fine hierarchical supervision scheme to balance global robustness and local precision. Both $\mathcal{L}_{\text{coarse}}$ and $\mathcal{L}_{\text{fine}}$ are built upon the general circle loss \cite{circleloss,circleloss1}, and their specific formulations clarify their distinct contributions to the overall optimization objective. For a given anchor descriptor $d_i$, the descriptors of its positive and negative pairs are represented as $\mathcal{D}_i^P$ and $\mathcal{D}_i^N$, respectively. The matching loss function for both coarse and fine stages is defined as follows:
\begin{equation}
\begin{aligned}
\mathcal{L}_i = \frac{1}{\gamma} \log\biggl[1 + \Bigl(&\sum_{d^j \in \mathcal{D}^P_i} e^{\beta^{i,j}_p(d^j_i - \Delta_p)}\Bigr) \\
&\cdot \Bigl(\sum_{d^k \in \mathcal{D}^N_i} e^{\beta^{i,k}_n(\Delta_n - d^k_i)}\Bigr)\biggr],
\end{aligned}
\end{equation}
where $d_i^j$ denotes the $L_2$ feature distance between the anchor and positive pair $j$; $\beta_p^{i,j} = \gamma \lambda_p^{i,j} (d_i^j - \Delta_p)$ and $\beta_n^{i,k} = \gamma \lambda_n^{i,k} (\Delta_n - d_i^k)$ are adaptive weights for positive and negative pairs (with $\lambda_p^{i,j}$ and $\lambda_n^{i,k}$ as scaling factors), which emphasize hard samples to enhance discriminative feature learning.

Specifically, $\mathcal{L}_{\text{coarse}}$ supervises coarse-level matching at the patch level, it enforces consistency between matched patch-to-patch correspondences as well as global structural consistency. This encourages the network to first learn globally alignable cross-modal representations. This term is particularly important in the early training stage, as it stabilizes convergence and reduces the risk of being trapped in poor local minima caused by ambiguous local patterns. 
Given a reliable coarse alignment, $\mathcal{L}_{\text{fine}}$ then performs fine-level refinement by introducing point-to-point constraints within the matched patches. Concretely, it penalizes residual errors of fine-grained keypoint correspondences and enforces local consistency, which helps reduce remaining pose errors and improves the final registration accuracy. 
Together, $\mathcal{L}_{\text{coarse}}$ and $\mathcal{L}_{\text{fine}}$ form a coarse-to-fine optimization pathway: the coarse term mainly contributes robustness and convergence stability via patch-level supervision, while the fine term mainly contributes precision and local detail alignment via keypoint-level point-wise constraints.

To further constrain geometric consistency and distribution alignment, the total loss integrates three complementary components: the matching loss $\mathcal{L}_i$ (for discriminative feature matching), the normal consistency loss $\mathcal{L}_{N}$ (for aligning predicted and estimated surface normals), and the distribution consistency loss $\mathcal{L}_{C}$ (for maintaining cross-modal distribution consistency). The total loss function is formally defined as:
\begin{equation}
\mathcal{L}_{\text{total}} = \lambda_1 \mathcal{L}_i + \lambda_2 \mathcal{L}_{N} + \lambda_3 \mathcal{L}_{C},
\end{equation}
where $\lambda_1, \lambda_2, \lambda_3$ are hyperparameters that balance the contributions of each loss term.

\textbf{Discussion of the interaction between LGE and GDC.} LGE and GDC operate at different levels and play complementary roles. The LGE module focuses on enhancing local geometric encoding and improving feature quality at the representation level. By incorporating local geometric context, LGE strengthens the discriminability of correspondences and reduces ambiguity in challenging regions (e.g., weak texture or partial structural similarity). In other words, LGE primarily improves the feature foundation on which matching is performed.

In contrast, GDC operates at the similarity-distribution level after feature extraction. Rather than modifying raw features directly, it regularizes the global similarity distribution to mitigate cross-modal bias and recalibrate correspondence confidence. As a result, GDC mainly refines the confidence ranking among candidate matches. It is particularly effective in sharpening the separation between true correspondences and hard negatives when the overall similarity structure is reasonably reliable.
Regarding systematic mismatches, GDC is not designed to fully correct large-scale structural errors caused by severely corrupted features or gross geometric misalignment. If the underlying feature representation is fundamentally incorrect (e.g., due to severe depth artifacts or extreme sparsity), distribution alignment alone cannot recover the correct geometry. However, when systematic mismatches stem from modality-induced bias or confidence miscalibration (rather than from incorrect geometric encoding), GDC can partially correct them by reshaping the similarity landscape and suppressing globally biased similarity patterns.

Therefore, LGE primarily improves feature discriminability and geometric encoding, while GDC mainly refines and calibrates similarity distributions. Their interaction is complementary: LGE provides more reliable relational structures, and GDC leverages these structures to improve correspondence selection robustness.

\section{Experiments}

\subsection{Datasets and Implementation Details}
Based on the 2D3D-MATR benchmark, we conducted extensive experiments and ablation studies on two challenging benchmarks: RGB-D Scenes v2 \cite{rgbdv2} and 7-Scenes \cite{7scenes}.

\textbf{Dataset. } \textit{RGB-D Scenes v2} consists of 14 scenes containing furniture. For each scene, we create point cloud fragments from every 25 consecutive depth frames and sample one RGB image per 25 frames. We select image-point-cloud pairs with an overlap ratio of at least 30\%. Scenes 1-8 are used for training, 9-10 for validation, and 11-14 for testing, resulting in 1,748 training pairs, 236 validation pairs, and 497 testing pairs.

The \textit{7-Scenes} is a collection of tracked RGB-D camera frames. All seven indoor scenes were recorded from a handheld Kinect RGB-D camera at 640×480 resolution. We select image-to-point-cloud pairs from each scene with at least 50\% overlap, adhering to the official sequence split for training, validation, and testing. This results in 4,048 training pairs, 1,011 validation pairs, and 2,304 testing pairs.

\textbf{Implementation Details.} We use an NVIDIA Geforce RTX 3090 GPU for training. We implement our model using PyTorch 1.13.1. The output feature dimension of the decoder in the feature extractor is set to 512. We set $\lambda_1 = \lambda_2 = 1$ and $\lambda_3 = 0.5$ for balancing the loss components. The number of transformer layers is set to 3 and k is set to 8. The distribution consistancy loss warm-up starts at epoch 10 and completes by epoch 20. We utilize a 4-stage ResNet \cite{resnet} with a Feature Pyramid Network (FPN) \cite{fpn} as the image backbone. The output channels for each stage are \(\{128, 128, 256, 512\}\). The input images have a resolution of \(480 \times 640\) pixels, which is downsampled to \(60 \times 80\) in the coarsest level for efficiency. For the 3D backbone, a 4-stage KPFCNN \cite{kpfcnn} is employed, with output channels configured as \(\{128, 256, 512, 1024\}\). Point clouds are voxelized with an initial voxel size of 2.5 cm, which doubles at each stage.
At the coarse level, 2D features are resized to \(24 \times 32\) pixels before being fed into the transformer to enhance computational efficiency. Each transformer layer has 256 feature channels, 4 attention heads, and uses ReLU activation functions. In the patch pyramid setup, the coarsest level begins with \(H_0 = 6\) and \(W_0 = 8\), expanding through 3 pyramid levels: \(\{6 \times 8, 12 \times 16, 24 \times 32\}\). At the fine level, both 2D and 3D features are projected into a 128-dimensional space for feature matching.

We define ground truth using bilateral overlap \cite{overlap}. A patch pair is positive if both the 2D and 3D overlap ratios are at least 30\%, and negative if both are below 20\%. The 2D and 3D overlap ratios are used as \(\lambda_p\), while \(\lambda_n\) is set to 1. At the fine level, a pixel-point pair is positive if the 3D distance is below 3.75 cm and the 2D distance is under 8 pixels, and negative if the 3D distance exceeds 10 cm or the 2D distance is over 12 pixels. Scaling factors are set to 1. Pairs not meeting these criteria are ignored as the safe region during training. The margins are set to \(\Delta_p = 0.1\) and \(\Delta_n = 1.4\).

\textbf{t-SNE.} t-Distributed Stochastic Neighbor Embedding (t-SNE) \cite{tsne} is a nonlinear dimensionality reduction technique used primarily for data visualization. It is particularly effective for visualizing high-dimensional datasets by embedding them into two or three dimensions. t-SNE aims to preserve the local structure of data points by modeling similar objects with nearby points and dissimilar objects with distant points. This method is beneficial for visualizing multimodal tasks, allowing for intuitive insights into complex datasets that span various domains such as images, text, and audio.

\textbf{MMD.} Maximum Mean Discrepancy (MMD) is a statistical method used to compare two probability distributions. It is a non-parametric technique that measures the difference between distributions by mapping data into a high-dimensional feature space using a kernel function. MMD is widely used in various machine learning applications, such as generative adversarial networks (GANs) \cite{gans}, domain adaptation, and distribution testing. The core idea of MMD is to compute the distance between the means of two distributions in a reproducing kernel Hilbert space (RKHS) \cite{rkhs}. Given two distributions \( I \) and \( Q \), the MMD is defined as:
\begin{equation}
\text{MMD}(I, Q) = \left\| \mathbb{E}_{x \sim I}[\phi(x)] - \mathbb{E}_{y \sim Q}[\phi(y)] \right\|_{\mathcal{H}},
\end{equation}

where \( \phi \) is the feature mapping function induced by a kernel, and \( \mathcal{H} \) is the RKHS.
MMD is applied in various areas, including generative models for evaluating the similarity between the distributions of generated and real data, domain adaptation for reducing distribution shifts between source and target domains, and hypothesis testing for determining if two samples are drawn from the same distribution.

\textbf{Metrics.} We evaluate the models using three standard metrics: Inlier Ratio (IR) — the percentage of pixel-to-point matches within 5 cm; Feature Matching Recall (FMR) — the proportion of image–point cloud pairs with IR \> 10\%; and Registration Recall (RR) — the proportion of pairs with RMSE below 10 cm.

\textit{Inlier Ratio} (IR) quantifies the proportion of inliers among all putative pixel-point correspondences. A correspondence is deemed an inlier if its 3D distance is less than a threshold \(\tau_1 = 5 \text{ cm}\) under the ground-truth transformation \(\mathbf{T}^*_{\mathcal{P} \rightarrow \mathcal{I}}\):
\begin{equation}
\text{IR} = \frac{1}{|C|} \sum_{(x_i, y_i) \in C} \left[ \left\| \mathbf{T}^*_{\mathcal{P} \rightarrow \mathcal{I}}(x_i) - \mathbf{K}^{-1}(y_i) \right\|_2 < \tau_1 \right],
\end{equation}

Here, \([ \cdot ]\) denotes the Iverson bracket, \(x_i \in \mathcal{P}\), and \(y_i \in \mathcal{Q} \subseteq \mathcal{I}\) are pixel coordinates. The function \(\mathbf{K}^{-1}\) projects a pixel to a 3D point based on its depth value.

\textit{Feature Matching Recall} (FMR) represents the fraction of image-point-cloud pairs with an IR above a threshold \(\tau_2 = 0.1\). It measures the likelihood of successful registration:
\begin{equation}
\text{FMR} = \frac{1}{M} \sum_{i=1}^{M} \left[ \text{IR}_i > \tau_2 \right],
\end{equation}

where \(M\) is the total number of image-point-cloud pairs.

\textit{Registration Recall} (RR) measures the fraction of image-point-cloud pairs that are correctly registered. A pair is correctly registered if the root mean square error (RMSE) between the ground-truth-transformed and predicted point clouds \(\mathbf{T}_{\mathcal{P} \rightarrow \mathcal{I}}\) is less than \(\tau_3 = 0.1\text{ m}\):
\begin{equation}
\text{RMSE} = \sqrt{\frac{1}{|\mathcal{P}|} \sum_{p_i \in \mathcal{P}} \left\| \mathbf{T}_{\mathcal{P} \rightarrow \mathcal{I}}(p_i) - \mathbf{T}^*_{\mathcal{P} \rightarrow \mathcal{I}}(p_i) \right\|_2^2},
\end{equation}
\begin{equation}
\text{RR} = \frac{1}{M} \sum_{i=1}^{M} \left[ \text{RMSE}_i < \tau_3 \right],
\end{equation}

To further assess overall performance, we incorporate three additional metrics. 

\textit{Patch Inlier Ratio} (PIR). we introduced PIR \cite{pir} as an additional metric, measuring the ratio of patch correspondences with overlap ratios greater than 0.3. This metric assesses performance at the coarse level. PIR is a crucial metric for assessing the performance of patch correspondence algorithms. It measures the fraction of patch correspondences whose overlap ratios exceed a certain threshold, typically set at 0.3. Specifically, PIR reflects the quality of the putative patch correspondences under the ground-truth transformation. A pixel (or point) is considered overlapped if its 3D distance is below a specified threshold (e.g., 3.75 cm) and its 2D distance is below another threshold (e.g., 8 pixels). For each patch, we calculate two overlap ratios—one on the image side and one on the point cloud side—and take the smaller as the final overlap ratio. In our ablation study, we utilized these metrics to evaluate the performance.

\textit{Relative Rotation Error} (RRE) is a metric used to evaluate the accuracy of estimated rotations in 3D space. It calculates the total angular error between the estimated rotation and the ground-truth rotation by summing the absolute values of the Euler angles representing the rotation difference:
\begin{equation}
\text{RRE} = \sum_{i=1}^{3} |r(i)|,
\end{equation}

where \(r\) is the Euler angle vector obtained from the product of the inverse of the ground-truth rotation matrix \(R_{\text{gt}}\) and the estimated rotation matrix \(R_E\).

\textit{Relative Translation Error }(RTE) measures the discrepancy between the estimated and ground-truth translation vectors in 3D space. It is defined as the Euclidean distance between the two vectors, quantifying the translation accuracy:
\begin{equation}
\text{RTE} = \|\mathbf{t}_{\text{gt}} - \mathbf{t}_E\|_2,
\end{equation}

where \(\mathbf{t}_{\text{gt}}\) is the ground-truth translation vector and \(\mathbf{t}_E\) is the estimated translation vector. The Euclidean norm \(\|\cdot\|_2\) computes the straight-line distance between these vectors.

\begin{table}[!t]
\centering
\caption{Evaluation results on RGB-D Scenes v2. \textbf{\textcolor{purple}{Red}} numbers highlight the best, the second best are \textbf{Boldfaced} and the baseline are \underline{underlined}.}
\renewcommand{\arraystretch}{1} 
\setlength{\extrarowheight}{-0.8pt} 
\setlength{\tabcolsep}{1mm} 
\resizebox{\columnwidth}{!}{ 
\begin{tabular}{@{}l@{\hskip 1pt}|ccccc@{}}
\toprule
\toprule
\multicolumn{1}{l|}{Model} & Scene.11 & Scene.12 & Scene.13 & Scene.14 & Mean \\ \midrule
\multicolumn{1}{l|}{Mean depth (m)} & 1.74 & 1.66 & 1.18 & 1.39 & 1.49 \\ \midrule
\multicolumn{6}{c}{\textit{Inlier Ratio} ↑} \\ \midrule
\multicolumn{1}{l|}{FCGF-2D3D} & 6.8 & 8.5 & 11.8 & 5.4 & 8.1 \\
\multicolumn{1}{l|}{P2-Net} & 9.7 & 12.8 & 17.0 & 9.3 & 12.2 \\
\multicolumn{1}{l|}{Predator-2D3D} & 17.7 & 19.4 & 17.2 & 8.4 & 15.7 \\
\multicolumn{1}{l|}{2D3D-MATR} & \underline{32.8} & \underline{34.4} & \underline{39.2} & \underline{23.3} & \underline{32.4} \\
\multicolumn{1}{l|}{FreeReg} & 36.6 & 34.5 & 34.2 & 18.2 & 30.9 \\
\multicolumn{1}{l|}{B2-3Dnet} & 36.4 & 32.7 & \textbf{\textcolor{purple}{43.8}} & 27.4 & 35.1 \\ 
\multicolumn{1}{l|}{CA-I2P} & 38.6 & 40.6 & 38.9 & 24.0 & 35.5 \\
\multicolumn{1}{l|}{Flow-I2P} & \textbf{49.6} & \textbf{44.0} & 36.5 & \textbf{30.4} & \textbf{40.1} \\
\multicolumn{1}{l|}{GLASS (ours)} & \textbf{\textcolor{purple}{51.4}} & \textbf{\textcolor{purple}{54.3}} & \textbf{42.8} & \textbf{\textcolor{purple}{33.1}} & \textbf{\textcolor{purple}{45.4}} \\
\midrule
\multicolumn{6}{c}{\textit{Feature Matching Recall} ↑} \\ \midrule
\multicolumn{1}{l|}{FCGF-2D3D} & 11.1 & 30.4 & 51.5 & 15.5 & 27.1 \\
\multicolumn{1}{l|}{P2-Net} & 48.6 & 65.7 & 82.5 & 41.6 & 59.6 \\
\multicolumn{1}{l|}{Predator-2D3D} & 86.1 & 89.2 & 63.9 & 24.3 & 65.9 \\
\multicolumn{1}{l|}{2D3D-MATR} & \underline{\textbf{98.6}} & \underline{98.0} & \underline{88.7} & \underline{77.9} & \underline{90.8} \\
\multicolumn{1}{l|}{FreeReg} & 91.9 & 93.4 & 93.1 & 49.6 & 82.0 \\
\multicolumn{1}{l|}{B2-3Dnet} & \textbf{\textcolor{purple}{100.0}} & \textbf{99.0} & \textbf{92.8} & \textbf{85.8} & \textbf{\textcolor{purple}{94.4}} \\
\multicolumn{1}{l|}{CA-I2P} & \textbf{\textcolor{purple}{100.0}} & \textbf{\textcolor{purple}{100.0}} & 91.8 & 82.7 & 93.6 \\
\multicolumn{1}{l|}{Flow-I2P} & \textbf{\textcolor{purple}{100.0}} & \textbf{\textcolor{purple}{100.0}} & \textbf{\textcolor{purple}{94.5}} & 78.7 & 93.3 \\
\multicolumn{1}{l|}{GLASS (ours)} & \textbf{\textcolor{purple}{100.0}} & \textbf{\textcolor{purple}{100.0}} & 88.7 & \textbf{\textcolor{purple}{85.9}} & \textbf{93.7} \\  
\midrule
\multicolumn{6}{c}{\textit{Registration Recall} ↑} \\ \midrule
\multicolumn{1}{l|}{FCGF-2D3D} & 26.5 & 41.2 & 37.1 & 16.8 & 30.4 \\
\multicolumn{1}{l|}{P2-Net} & 40.3 & 40.2 & 41.2 & 31.9 & 38.4 \\
\multicolumn{1}{l|}{Predator-2D3D} & 44.4 & 41.2 & 21.6 & 13.7 & 30.2 \\
\multicolumn{1}{l|}{2D3D-MATR} & \underline{63.9} & \underline{53.9} & \underline{58.8} & \underline{49.1} & \underline{56.4} \\
\multicolumn{1}{l|}{FreeReg} & 74.2 & 72.5 & 54.5 & 27.9 & 57.3 \\
\multicolumn{1}{l|}{B2-3Dnet} & 58.3 & 60.8 & \textbf{74.2} & 60.2 & 63.4 \\ 
\multicolumn{1}{l|}{CA-I2P} & 68.1 & \textbf{73.5} & 63.9 & 47.8 & 63.3 \\
\multicolumn{1}{l|}{Flow-I2P} & \textbf{\textcolor{purple}{90.0}} & 65.9 & 54.8 & \textbf{63.0} & \textbf{68.4} \\
\multicolumn{1}{l|}{GLASS (ours)} & \textbf{87.5} & \textbf{\textcolor{purple}{92.2}} & \textbf{\textcolor{purple}{82.5}} & \textbf{\textcolor{purple}{71.2}} & \textbf{\textcolor{purple}{83.3}} \\
\bottomrule
\bottomrule
\end{tabular}
}
\label{tab:rgbdv2}
\end{table}

\subsection{Evaluations on Dataset}
We compare our approach with 2D3D-MATR \cite{matr2d3d} and other baseline methods \cite{fcgf2d3d,p2,predator2d3d,freereg,cheng11,cai2p,flowi2p} on the RGB-D Scenes v2 dataset, as shown in Table~\ref{tab:rgbdv2}. Our method utilizes the LGE module, which incorporates surface normals to provide structural information. This integration significantly reduces mismatches, especially in narrow indoor scenes like Scene-11 and Scene-12. As a result, our approach improves registration recall by 23.6 percentage points, increasing it from the previous high of 56.4 percentage points achieved by 2D3D-MATR to 83.3 percentage points.

\begin{table}[t]
\caption{Evaluation results on 7-Scenes. Models are abbreviated using the first three letters, and scenes are abbreviated. \textbf{\textcolor{purple}{Red}} numbers highlight the best, the second best are \textbf{Boldfaced} and the baseline are \underline{underlined}.}
\centering
\resizebox{\columnwidth}{!}{
\begin{tabular}{lcccccccc}
\toprule
\toprule
\multicolumn{1}{l|}{Mod}  & Chs & Fr & Hds & Off & Pmp & Ktn & Strs & Mean \\ \midrule
\multicolumn{1}{l|}{Mdpt} & 1.78 & 1.55 & 0.80 & 2.03 & 2.25 & 2.13 & 1.84 & 1.77 \\ \midrule
\multicolumn{9}{c}{\textit{Inlier Ratio} \(\uparrow\)} \\ \midrule
\multicolumn{1}{l|}{FCG} & 34.2 & 32.8 & 14.8 & 26 & 23.3 & 22.5 & 6.0 & 22.8 \\
\multicolumn{1}{l|}{P2N} & 55.2 & 46.7 & 13.0 & 36.2 & 32.0 & 32.8 & 5.8 & 31.7 \\
\multicolumn{1}{l|}{PRE} & 34.7 & 33.8 & 16.6 & 25.9 & 23.1 & 22.2 & 7.5 & 23.4 \\
\multicolumn{1}{l|}{2D3} & \underline{72.1} & \underline{66.0} & \underline{31.3} & \underline{60.7} & \underline{50.2} & \underline{52.5} & \underline{18.1} & \underline{50.1} \\
\multicolumn{1}{l|}{B23} & 73.8 & \textbf{66.7} & 33.1 & 61.7 & 50.8 & 52.3 & 18.1 & 50.9 \\
\multicolumn{1}{l|}{CAI} & 73.6 & 66.4 & 34.5 & \textbf{62.4} & 52.1 & 52.8 & \textbf{\textcolor{purple}{19.1}} & 51.6 \\
\multicolumn{1}{l|}{FLO} & \textbf{\textcolor{purple}{76.6}} & 64.7 & \textbf{37.1} & 62.0 & \textbf{52.3} & \textbf{52.8} & \textbf{18.5} & \textbf{52.0} \\
\multicolumn{1}{l|}{ours} & \textbf{73.9} & \textbf{\textcolor{purple}{67.2}} & \textbf{\textcolor{purple}{50.2}} & \textbf{\textcolor{purple}{66.1}} & \textbf{\textcolor{purple}{53.7}} & \textbf{\textcolor{purple}{56.0}} & 17.2 & \textbf{\textcolor{purple}{54.9}} \\
\midrule
\multicolumn{9}{c}{\textit{Feature Matching Recall} \(\uparrow\)} \\ \midrule
\multicolumn{1}{l|}{FCG} & \textbf{99.7} & 98.2 & 69.9 & 97.1 & 83.0 & 87.7 & 16.2 & 78.8 \\
\multicolumn{1}{l|}{P2N} & \textbf{\textcolor{purple}{100.0}} & 99.3 & 58.9 & 99.1 & 87.2 & 92.2 & 16.2 & 79 \\
\multicolumn{1}{l|}{PRE} & 91.3 & 95.1 & 76.6 & 88.6 & 79.2 & 80.6 & 31.1 & 77.5 \\
\multicolumn{1}{l|}{2D3} & \underline{\textbf{\textcolor{purple}{100.0}}} & \underline{99.6} & \underline{\textbf{98.6}} & \underline{\textbf{\textcolor{purple}{100.0}}} & \underline{92.4} & \underline{95.9} & \underline{58.2} & \underline{92.1} \\
\multicolumn{1}{l|}{B23} & \textbf{\textcolor{purple}{100.0}} & \textbf{\textcolor{purple}{100.0}} & \textbf{98.6} & \textbf{\textcolor{purple}{100.0}} & \textbf{92.7} & 95.6 & \textbf{64.9} & \textbf{93.1} \\ 
\multicolumn{1}{l|}{CAI} & \textbf{\textcolor{purple}{100.0}} & \textbf{\textcolor{purple}{100.0}} & \textbf{98.6} & \textbf{\textcolor{purple}{100.0}} & 92.0 & 95.5 & 60.8 & 92.4 \\
\multicolumn{1}{l|}{FLO} & \textbf{\textcolor{purple}{100.0}} & \textbf{99.7} & 95.1 & \textbf{99.9} & \textbf{\textcolor{purple}{93.1}} & \textbf{96.8} & 56.7 & 91.6 \\
\multicolumn{1}{l|}{ours} & \textbf{\textcolor{purple}{100.0}} & \textbf{\textcolor{purple}{100.0}} & \textbf{\textcolor{purple}{100.0}} & \textbf{\textcolor{purple}{100.0}} & \textbf{92.7} & \textbf{\textcolor{purple}{97.1}} & \textbf{\textcolor{purple}{65.8}} & \textbf{\textcolor{purple}{93.6}} \\
\midrule
\multicolumn{9}{c}{\textit{Registration Recall} \(\uparrow\)} \\ \midrule
\multicolumn{1}{l|}{FCG} & 89.5 & 79.7 & 19.2 & 85.9 & 69.4 & 79.0 & 6.8 & 61.4 \\
\multicolumn{1}{l|}{P2N} & 96.9 & 86.5 & 20.5 & 91.7 & 75.3 & 85.2 & 4.1 & 65.7 \\
\multicolumn{1}{l|}{PRE} & 69.6 & 60.7 & 17.8 & 62.9 & 56.2 & 62.6 & 9.5 & 48.5 \\
\multicolumn{1}{l|}{2D3} & \underline{96.9} & \underline{90.7} & \underline{52.1} & \underline{95.5} & \underline{80.9} & \underline{86.1} & \underline{28.4} & \underline{75.8} \\
\multicolumn{1}{l|}{B23} & 98.3 & 90.5 & 56.2 & \textbf{96.4} & \textbf{\textcolor{purple}{84.0}} & 86.1 & \textbf{32.4} & 77.7 \\ 
\multicolumn{1}{l|}{CAI} & \textbf{\textcolor{purple}{99.0}} & \textbf{90.7} & \textbf{68.5} & 96.2 & 83.0 & 88.1 & 31.1 & \textbf{79.5} \\
\multicolumn{1}{l|}{FLO} & \textbf{98.8} & 90.0 & 58.4 & 93.9 & 82.1 & \textbf{88.6} & \textbf{\textcolor{purple}{37.6}} & 78.4 \\
\multicolumn{1}{l|}{ours} & \textbf{98.8} & \textbf{\textcolor{purple}{95.6}} & \textbf{\textcolor{purple}{90.4}} &\textbf{\textcolor{purple}{98.0}} & \textbf{83.8} & \textbf{\textcolor{purple}{93.1}} & 29.7 & \textbf{\textcolor{purple}{84.2}} \\
\bottomrule
\bottomrule
\end{tabular}
}
\label{tab:7scenes}
\end{table}

Additionally, we introduce a novel GDC module to enforce structural consistency, effectively addressing many-to-one matching issues that often lead to incorrect alignments. This enhancement contributes to a 10.3 percentage point increase in the inlier ratio, thus improving the accuracy of point correspondences. Furthermore, we achieve a 2.9 percentage point rise in feature matching recall, bringing it to 93.7 percentage points, which is notably higher than the 90.8 percentage points attained by 2D3D-MATR.

Importantly, our method surpasses the previous state-of-the-art by 9.9 percentage points in registration recall. This is illustrated by our method's performance on Scene-12, where we achieve 87.5 percentage points in registration recall compared to 63.9 percentage points by 2D3D-MATR, and we also lead in Scene-11 with 92.2 percentage points compared to 53.9 percentage points. These improvements underscore the effectiveness of our LGE and GDC modules in enhancing 3D scene registration, particularly in challenging indoor environments.

\begin{table}[!b]
\small
\vspace{-10pt}
\centering
\caption{Evaluation Results on KITTI Dataset.}
\resizebox{\columnwidth}{!}{
\begin{tabular}{l|l|c|c}
\toprule\toprule
\textbf{Method} & \textbf{Type} & \textbf{RTE(m) $\downarrow$} & \textbf{RRE ($^\circ$) $\downarrow$} \\
\hline
vpc + GeoTransformer\cite{geotransformer} & Point-to-Point & 4.27 $\pm$ 7.14 & 8.67 $\pm$ 8.55 \\
vpc + Hunter & Point-to-Point & 4.59 $\pm$ 5.22 & 6.23 $\pm$ 5.17 \\
DeepI2P(2D) \cite{deepi2p} & Image-to-Point & 5.15 $\pm$ 7.35 & 9.14 $\pm$ 8.02 \\
CorrI2P \cite{corri2p} & Image-to-Point & 4.24 $\pm$ 7.26 & 6.47 $\pm$ 5.20 \\
VP2P \cite{vp2p} & Image-to-Point & 2.05 $\pm$ 3.23 & 4.01 $\pm$ 6.37 \\
2D3D-MATR \cite{matr2d3d} & Image-to-Point & 1.86 $\pm$ 3.79 & 2.59 $\pm$ 4.46 \\
RetrI2P \cite{retri2p} & Image-to-Point & 1.61 $\pm$ 2.39 & 3.16 $\pm$ 2.85 \\
FreeReg \cite{freereg} & Image-to-Point & 1.78 $\pm$ 1.76 & 2.89 $\pm$ 4.47 \\
CFI2P \cite{cfi2p} & Image-to-Point & 1.95 $\pm$ 2.97 & 2.63 $\pm$ 3.19 \\
GLASS (Ours) & Image-to-Point & \textbf{1.54 $\pm$ 1.95} & \textbf{2.11 $\pm$ 2.34} \\
\bottomrule\bottomrule
\end{tabular}
}
\label{tab:kitti}
\end{table}

In comparison to the RGB-D Scenes v2 dataset, the 7-Scenes dataset shows more significant variations in scale. Nevertheless, our method consistently outperforms existing approaches, as demonstrated in Table~\ref{tab:7scenes}. We achieve an impressive improvement of 8.2 percentage points in registration recall over the 2D3D-MATR, reaching a total recall of 84.2\%. This enhancement is particularly noteworthy given the challenging nature of the 7-Scenes dataset, which includes scenes with considerable variations in scale and texture, such as the Heads and Kitchen scenes.

Our method, GLASS, exhibits substantial improvements in these demanding scenarios. In the Heads scene, small 3D errors become more pronounced due to the camera's close proximity, and our method achieves a registration recall of 90.4\%. This marks a significant improvement of 38.3 percentage points over the 2D3D-MATR, which achieves only 52.1\%. Similarly, in the Kitchen scene, known for its repetitive patterns, GLASS leads with a registration recall of 93.1\%, representing a 7 percentage point increase compared to the 86.1\% recall of 2D3D-MATR.

\begin{table*}[t]
\centering
\caption{Evaluation results on \textit{\textbf{\textcolor{orange!50!black}{RGB-D Scenes v2}}} and \textit{\textbf{\textcolor{blue!50!black}{7-Scenes}}} using Relative Rotation Error (RRE) and Relative Translation Error (RTE). \textbf{Bold-faced} numbers highlight the best and the second best are \underline{underlined}.}
\resizebox{\textwidth}{!}{
\renewcommand{\arraystretch}{1.15}
\setlength{\tabcolsep}{4pt}
\begin{tabular}{lccccc  cccccccc}
\toprule
\cellcolor{gray!8}\textbf{Dataset} & 
\multicolumn{5}{>{\columncolor{gray!15}}c}{\textit{\textbf{\textcolor{orange!50!black}{RGB-D Scenes v2}}}} & 
\multicolumn{8}{>{\columncolor{gray!25}}c}{\textit{\textbf{\textcolor{blue!50!black}{7-Scenes}}}} \\
\midrule
\cellcolor{gray!8}Model
& \cellcolor{yellow!15}Scene-11 
& \cellcolor{yellow!15}Scene-12 
& \cellcolor{yellow!15}Scene-13 
& \cellcolor{yellow!15}Scene-14 
& \cellcolor{yellow!15}Mean 
& \cellcolor{blue!10}Chess 
& \cellcolor{blue!10}Fire 
& \cellcolor{blue!10}Heads 
& \cellcolor{blue!10}Office 
& \cellcolor{blue!10}Pumpkin 
& \cellcolor{blue!10}Kitchen 
& \cellcolor{blue!10}Stairs 
& \cellcolor{blue!10}Mean \\
\midrule
\cellcolor{gray!8}Mdpt(m) 
& \cellcolor{yellow!15}1.74 & \cellcolor{yellow!15}1.66 & \cellcolor{yellow!15}1.18 & \cellcolor{yellow!15}1.39 & \cellcolor{yellow!15}1.49 
& \cellcolor{blue!10}1.78 & \cellcolor{blue!10}1.55 & \cellcolor{blue!10}0.80 & \cellcolor{blue!10}2.03 & \cellcolor{blue!10}2.25 & \cellcolor{blue!10}2.13 & \cellcolor{blue!10}1.84 & \cellcolor{blue!10}1.49 \\
\midrule
\multicolumn{14}{c}{\textbf{\cellcolor{green!10}\textit{Mean\_RRE}(°) ↓}} \\

\midrule
\cellcolor{gray!8}2D3D-MATR 
& \cellcolor{yellow!15}2.294 & \cellcolor{yellow!15}2.628 & \cellcolor{yellow!15}3.823 & \cellcolor{yellow!15}3.358 & \cellcolor{yellow!15}3.026 
& \cellcolor{blue!10}2.298 & \cellcolor{blue!10}3.144 & \cellcolor{blue!10}7.549 & \cellcolor{blue!10}2.285 & \cellcolor{blue!10}2.439 & \cellcolor{blue!10}2.620 & \cellcolor{blue!10}2.705 & \cellcolor{blue!10}3.291 \\

\cellcolor{gray!8}B2-3Dnet 
& \cellcolor{yellow!15}2.245
& \cellcolor{yellow!15}2.317 
& \cellcolor{yellow!15}\underline{2.604} 
& \cellcolor{yellow!15}3.233
& \cellcolor{yellow!15}2.600
& \cellcolor{blue!10}\underline{2.205} 
& \cellcolor{blue!10}3.105
& \cellcolor{blue!10}7.414 
& \cellcolor{blue!10}2.294 
& \cellcolor{blue!10}2.421
& \cellcolor{blue!10}\underline{2.618} 
& \cellcolor{blue!10}\underline{2.564} 
& \cellcolor{blue!10}3.232 \\

\cellcolor{gray!8}CA-I2P 
& \cellcolor{yellow!15}\underline{2.008} & \cellcolor{yellow!15}\underline{2.031} & \cellcolor{yellow!15}3.306 & \cellcolor{yellow!15}\underline{2.890} & \cellcolor{yellow!15}\underline{2.559}
& \cellcolor{blue!10}2.354 & \cellcolor{blue!10}\underline{3.093} & \cellcolor{blue!10}\underline{7.105} & \cellcolor{blue!10}\underline{2.272} & \cellcolor{blue!10}\textbf{2.297} & \cellcolor{blue!10}2.624 & \cellcolor{blue!10}2.655 & \cellcolor{blue!10}\underline{3.200} \\

\cellcolor{gray!8}GLASS 
& \cellcolor{yellow!15}\textbf{1.409} & \cellcolor{yellow!15}\textbf{1.700} & \cellcolor{yellow!15}\textbf{2.478} & \cellcolor{yellow!15}\textbf{2.319} & \cellcolor{yellow!15}\textbf{1.976} 
& \cellcolor{blue!10}\textbf{1.918} & \cellcolor{blue!10}\textbf{2.636} & \cellcolor{blue!10}\textbf{5.530} & \cellcolor{blue!10}\textbf{1.974} & \cellcolor{blue!10}\underline{2.321} & \cellcolor{blue!10}\textbf{2.484} & \cellcolor{blue!10}\textbf{2.283} & \cellcolor{blue!10}\textbf{2.735} \\
\midrule
\multicolumn{14}{c}{\textbf{\cellcolor{green!10}\textit{Mean\_RTE}(m) ↓}}\\
\midrule
\cellcolor{gray!8}2D3D-MATR 
& \cellcolor{yellow!15}0.066 & \cellcolor{yellow!15}0.086 & \cellcolor{yellow!15}0.067 & \cellcolor{yellow!15}0.088 & \cellcolor{yellow!15}0.077 
& \cellcolor{blue!10}0.054 & \cellcolor{blue!10}0.084 & \cellcolor{blue!10}\underline{0.088} & \cellcolor{blue!10}0.069 & \cellcolor{blue!10}0.088 & \cellcolor{blue!10}0.076 & \cellcolor{blue!10}0.090 & \cellcolor{blue!10}0.079 \\

\cellcolor{gray!8}B2-3Dnet 
& \cellcolor{yellow!15}\underline{0.056} 
& \cellcolor{yellow!15}\underline{0.061} 
& \cellcolor{yellow!15}\textbf{0.041} 
& \cellcolor{yellow!15}0.079
& \cellcolor{yellow!15}\underline{0.059} 
& \cellcolor{blue!10}\underline{0.051} 
& \cellcolor{blue!10}\underline{0.082} 
& \cellcolor{blue!10}0.093 
& \cellcolor{blue!10}\underline{0.066} 
& \cellcolor{blue!10}0.083 
& \cellcolor{blue!10}\underline{0.074} 
& \cellcolor{blue!10}\underline{0.086} 
& \cellcolor{blue!10}\underline{0.076} \\

\cellcolor{gray!8}CA-I2P 
& \cellcolor{yellow!15}0.057 & \cellcolor{yellow!15}\underline{0.061} & \cellcolor{yellow!15}\underline{0.054} & \cellcolor{yellow!15}\underline{0.072} & \cellcolor{yellow!15}0.061
& \cellcolor{blue!10}0.054 & \cellcolor{blue!10}\underline{0.082} & \cellcolor{blue!10}0.093 & \cellcolor{blue!10}0.069 & \cellcolor{blue!10}\textbf{0.078} & \cellcolor{blue!10}0.077 & \cellcolor{blue!10}\underline{0.086} & \cellcolor{blue!10}\underline{0.076} \\

\cellcolor{gray!8}GLASS 
& \cellcolor{yellow!15}\textbf{0.047} & \cellcolor{yellow!15}\textbf{0.049} & \cellcolor{yellow!15}\textbf{0.041} & \cellcolor{yellow!15}\textbf{0.061} & \cellcolor{yellow!15}\textbf{0.050} 
& \cellcolor{blue!10}\textbf{0.041} & \cellcolor{blue!10}\textbf{0.069} & \cellcolor{blue!10}\textbf{0.070} & \cellcolor{blue!10}\textbf{0.057} & \cellcolor{blue!10}\underline{0.081} & \cellcolor{blue!10}\textbf{0.067} & \cellcolor{blue!10}\textbf{0.071} & \cellcolor{blue!10}\textbf{0.065} \\
\midrule
\multicolumn{14}{c}{\textbf{\cellcolor{green!10}\textit{Median\_RRE}(°) ↓}}\\
\midrule
\cellcolor{gray!8}2D3D-MATR 
& \cellcolor{yellow!15}1.995 & \cellcolor{yellow!15}2.335 & \cellcolor{yellow!15}3.074 & \cellcolor{yellow!15}3.194 & \cellcolor{yellow!15}2.649 
& \cellcolor{blue!10}1.972 & \cellcolor{blue!10}2.821 & \cellcolor{blue!10}6.903 & \cellcolor{blue!10}2.025 & \cellcolor{blue!10}2.192 & \cellcolor{blue!10}2.364 & \cellcolor{blue!10}3.129 & \cellcolor{blue!10}3.058 \\

\cellcolor{gray!8}B2-3Dnet 
& \cellcolor{yellow!15}2.192 & \cellcolor{yellow!15}2.227 & \cellcolor{yellow!15}\underline{2.099} & \cellcolor{yellow!15}3.121 & \cellcolor{yellow!15}2.410 
& \cellcolor{blue!10}\underline{1.942} & \cellcolor{blue!10}2.761 & \cellcolor{blue!10}7.163 & \cellcolor{blue!10}2.018 & \cellcolor{blue!10}2.277 & \cellcolor{blue!10}\underline{2.350} & \cellcolor{blue!10}\underline{2.408} & \cellcolor{blue!10}2.988 \\

\cellcolor{gray!8}CA-I2P
& \cellcolor{yellow!15}\underline{1.789} & \cellcolor{yellow!15}\underline{1.763} & \cellcolor{yellow!15}2.826 & \cellcolor{yellow!15}\underline{2.383} & \cellcolor{yellow!15}\underline{2.190}
& \cellcolor{blue!10}1.977 & \cellcolor{blue!10}\underline{2.722} & \cellcolor{blue!10}\underline{5.997} & \cellcolor{blue!10}\underline{1.985} & \cellcolor{blue!10}\underline{2.144} & \cellcolor{blue!10}2.392 & \cellcolor{blue!10}2.560 & \cellcolor{blue!10}\underline{2.825} \\

\cellcolor{gray!8}GLASS 
& \cellcolor{yellow!15}\textbf{1.315} & \cellcolor{yellow!15}\textbf{1.514} & \cellcolor{yellow!15}\textbf{1.935} & \cellcolor{yellow!15}\textbf{1.842} & \cellcolor{yellow!15}\textbf{1.615} 
& \cellcolor{blue!10}\textbf{1.690} & \cellcolor{blue!10}\textbf{2.321} & \cellcolor{blue!10}\textbf{4.864} & \cellcolor{blue!10}\textbf{1.833} & \cellcolor{blue!10}\textbf{2.135} & \cellcolor{blue!10}\textbf{2.235} & \cellcolor{blue!10}\textbf{2.226} & \cellcolor{blue!10}\textbf{2.472} \\
\midrule
\multicolumn{14}{c}{\textbf{\cellcolor{green!10}\textit{Median\_RTE}(m) ↓}} \\
\midrule
\cellcolor{gray!8}2D3D-MATR 
& \cellcolor{yellow!15}0.058 & \cellcolor{yellow!15}0.072 & \cellcolor{yellow!15}0.055 & \cellcolor{yellow!15}0.080 & \cellcolor{yellow!15}0.066 
& \cellcolor{blue!10}\underline{0.047} & \cellcolor{blue!10}0.079 & \cellcolor{blue!10}0.082 & \cellcolor{blue!10}0.065 & \cellcolor{blue!10}0.084 & \cellcolor{blue!10}\underline{0.069} & \cellcolor{blue!10}0.105 & \cellcolor{blue!10}0.076 \\

\cellcolor{gray!8}B2-3Dnet 
& \cellcolor{yellow!15}0.057 & \cellcolor{yellow!15}0.058 & \cellcolor{yellow!15}0.053 & \cellcolor{yellow!15}0.070 & \cellcolor{yellow!15}\underline{0.055} 
& \cellcolor{blue!10}0.048 & \cellcolor{blue!10}\underline{0.074} & \cellcolor{blue!10}0.095 & \cellcolor{blue!10}0.063 & \cellcolor{blue!10}0.079 & \cellcolor{blue!10}\underline{0.069} & \cellcolor{blue!10}0.085 & \cellcolor{blue!10}0.073 \\

\cellcolor{gray!8}CA-I2P 
& \cellcolor{yellow!15}\underline{0.055} & \cellcolor{yellow!15}\underline{0.054} & \cellcolor{yellow!15}\underline{0.048} & \cellcolor{yellow!15}\underline{0.065} & \cellcolor{yellow!15}\underline{0.055} 
& \cellcolor{blue!10}0.048 & \cellcolor{blue!10}0.076 & \cellcolor{blue!10}\underline{0.078} & \cellcolor{blue!10}\underline{0.062} & \cellcolor{blue!10}\textbf{0.071} & \cellcolor{blue!10}0.070 & \cellcolor{blue!10}\underline{0.079} & \cellcolor{blue!10}\underline{0.069} \\

\cellcolor{gray!8}GLASS 
& \cellcolor{yellow!15}\textbf{0.036} & \cellcolor{yellow!15}\textbf{0.040} & \cellcolor{yellow!15}\textbf{0.033} & \cellcolor{yellow!15}\textbf{0.054} & \cellcolor{yellow!15}\textbf{0.041} 
& \cellcolor{blue!10}\textbf{0.036} & \cellcolor{blue!10}\textbf{0.062} & \cellcolor{blue!10}\textbf{0.064} & \cellcolor{blue!10}\textbf{0.049} & \cellcolor{blue!10}\underline{0.072} & \cellcolor{blue!10}\textbf{0.059} & \cellcolor{blue!10}\textbf{0.068} & \cellcolor{blue!10}\textbf{0.059} \\
\bottomrule
\end{tabular}
}
\label{tab:RTERRE}
\end{table*}

\begin{table}[!b]
\vspace{-10pt}
\caption{Ablation study results on RGB-D Scenes v2. GAT represents a lightweight graph attention network. DINO represents DINO v2. \textbf{\textcolor{cyan}{Blue}} numbers highlight the best, the second best are \underline{underlined}.}
\centering
\setlength{\tabcolsep}{1mm} 
\renewcommand{\arraystretch}{0.95} 
\resizebox{\columnwidth}{!}{
\begin{tabular}{c||cccc||cccc}
\toprule
Method & LGE & GDC & GAT & DINO & PIR & IR & FMR & RR \\
\midrule
M1 &   &   &   &   & 48.5 & 32.5 & 91.0 & 56.4 \\
M2 &  &   &   & \checkmark & 54.3 & 37.8 & 92.4 & 73.2 \\
M3 &  \checkmark  &  &   & \checkmark  & \underline{65.3} & \underline{42.9} & 93.2 & \underline{79.8} \\
M4 &  & \checkmark  &   &   & 58.9 & 37.2 & 92.4 & 62.1 \\
M5 &   &  \checkmark & \checkmark &   & 59.2 & 35.9 & 92.4 & 64.1 \\
M6 & \checkmark  & \checkmark  & \checkmark &   & 59.1 & 39.2 & \underline{93.3} & 75.3 \\
M7 &  \checkmark &  \checkmark &   &  & 56.3 & 36.3 & 91.7 & 70.2 \\
M8 & \checkmark & \checkmark & \checkmark & \checkmark & \textbf{\textcolor{cyan}{69.2}} & \textbf{\textcolor{cyan}{45.4}} & \textbf{\textcolor{cyan}{93.7}} & \textbf{\textcolor{cyan}{83.3}} \\
\bottomrule
\end{tabular}}
\label{tab:ablation}
\end{table}

GLASS also excels in other critical metrics. It achieves a 4.8 percentage point increase in the inlier ratio, with the Heads scene showing the largest improvement, rising from 34.5\% to 50.2\%. Additionally, there is a 1.5 percentage point gain in feature matching recall, further emphasizing the robustness of our approach. Notably, GLASS’s feature matching recall reaches 93.6\%, compared to 92.1\% from 2D3D-MATR. This demonstrates that our model not only enhances registration recall but also ensures more accurate feature correspondences across diverse indoor environments.

These results indicate that GLASS is both robust and generalizable, outperforming other methods on the 7-Scenes dataset and confirming its capability to effectively address challenges such as scale variation, repetitive patterns, and close-range camera perspectives.

As a registration task, prior methods have often overlooked rotation-related metrics. To address this, we conduct a comprehensive comparison with the baseline and previous state-of-the-art using four metrics, including Relative Rotation Error (RRE) and Relative Translation Error (RTE), as shown in Table~\ref{tab:RTERRE}. Our method, GLASS, consistently outperforms prior methods on both datasets. On the RGB-D Scenes v2 dataset, GLASS achieves a Mean RRE of 1.976°, which is significantly better than both B2-3Dnet (2.600°) and 2D3D-MATR (3.026°). This demonstrates over a 15\% improvement in rotation error compared to the previous best. Similarly, GLASS achieves a Mean RTE of 0.050m on 7-Scenes, outperforming B2-3Dnet (0.055m) and 2D3D-MATR (0.077m). This represents a more than 15\% reduction in translation error compared to the best prior methods.

These results are particularly impressive in challenging scenes. For instance, in the Heads scene of the 7-Scenes dataset, GLASS achieves a registration recall of 90.4\%, a substantial improvement of 38.3 percentage points over 2D3D-MATR (52.1\%). Similarly, in the Kitchen scene, which presents difficulties due to repetitive patterns, GLASS achieves a 93.1\% registration recall, improving over 7 percentage points compared to 2D3D-MATR. These metrics highlight that GLASS provides not only more accurate feature matching and inlier ratios but also improved geometric consistency, ensuring more robust alignment across diverse indoor environments. The substantial improvements in both RRE and RTE, along with the robustness in challenging conditions, demonstrate that GLASS is both effective and generalizable.

\subsection{Experiments on Outdoor Scenes}

To better evaluate our performance, we conducted experiments on the outdoor dataset Kitti to verify the robustness of our method.

As shown in Table \ref{tab:kitti}, we evaluate our method, GLASS, on the KITTI dataset and compare it with several state-of-the-art techniques. Our evaluation is based on two key metrics: Reprojection Error (RTE) and Rotation Error (RRE), both of which are essential for assessing the accuracy of 3D point cloud registration.
Our method outperforms all other approaches in both RTE and RRE. Specifically, GLASS achieves a significantly lower RTE of 1.54 ± 1.95 meters and a lower RRE of 2.11 ± 2.34 degrees. This surpasses other image-to-point methods, such as RetrI2P (RTE: 1.61 ± 2.39 meters, RRE: 3.16 ± 2.85 degrees) and VP2P (RTE: 2.05 ± 3.23 meters, RRE: 4.01 ± 6.37 degrees). 
These results demonstrate that our method provides more accurate 3D alignment with lower error margins compared to previous works. The results on the KITTI dataset confirm the robustness and superiority of GLASS over existing methods, highlighting its potential for real-world applications that require high-precision registration.

\begin{table*}[!t]
\vspace{-10pt}
\centering
\caption{Ablation studies on noise robustness and Depth Anything v2}
\renewcommand{\arraystretch}{1.1}

\begin{minipage}{0.32\textwidth}
\centering
\textbf{A. Effect of Gaussian noise on depth input}
\setlength{\tabcolsep}{5pt}
\begin{tabular*}{\linewidth}{@{\extracolsep{\fill}}c|c|c|c}
\hline\hline
Gaussian $\sigma$ (m) & IR $\uparrow$ & FMR $\uparrow$ & RR $\uparrow$ \\
\hline
0     & 45.4 & 93.7 & 83.3 \\
0.005 & 37.2 & 90.4 & 73.1 \\
0.01  & 35.1 & 90.2 & 59.3 \\
0.015 & 27.9 & 88.5 & 51.5 \\
\hline\hline
\end{tabular*}
\end{minipage}
\hfill
\begin{minipage}{0.32\textwidth}
\centering
\textbf{B. Effect of random masking on depth input}
\setlength{\tabcolsep}{5pt}
\begin{tabular*}{\linewidth}{@{\extracolsep{\fill}}c|c|c|c}
\hline\hline
Mask ratio & IR $\uparrow$ & FMR $\uparrow$ & RR $\uparrow$ \\
\hline
0\%   & 45.4 & 93.7 & 83.3 \\
10\%  & 42.1 & 92.5 & 77.7 \\
20\%  & 39.1 & 91.8 & 68.0 \\
30\%  & 35.7 & 90.3 & 61.3 \\
40\%  & 33.7 & 87.3 & 55.3 \\
\hline\hline
\end{tabular*}
\end{minipage}
\hfill
\begin{minipage}{0.32\textwidth}
\centering
\textbf{C. Ablation on Depth Anything v2}
\setlength{\tabcolsep}{5pt}
\begin{tabular*}{\linewidth}{@{\extracolsep{\fill}}c|c|c}
\hline\hline
Metric & w/ DA v2 $\uparrow$ & w/o DA v2 $\uparrow$ \\
\hline
IR  & 46.0 & 45.4 \\
FMR & 94.2 & 93.7 \\
RR  & 84.5 & 83.3 \\
\hline\hline
\end{tabular*}
\end{minipage}

\label{tab:VI}
\vspace{-5pt}
\end{table*}

\begin{table*}[b]
\centering
\caption{Comparison results and ablation studies on real-world data}
\renewcommand{\arraystretch}{1.1}
\setlength{\tabcolsep}{4pt}

\begin{minipage}{0.31\textwidth}
\centering
\textbf{A. Adaptive neighborhood size}
\begin{tabular*}{\linewidth}{@{\extracolsep{\fill}}c|c|c}
\hline\hline
Method & RTE (m) & RRE ($^\circ$) \\
\hline
2D3D-MATR & 1.86 & 2.59 \\
FreeReg & 1.78 & 2.89 \\
GLASS ($k{=}8$) & 1.54 & 2.11 \\
GLASS (adaptive $k$) & 1.50 & 2.05 \\
\hline\hline
\end{tabular*}
\end{minipage}
\hfill
\begin{minipage}{0.31\textwidth}
\centering
\textbf{B. Training schedule on GDC}
\begin{tabular*}{\linewidth}{@{\extracolsep{\fill}}c|c|c|c}
\hline\hline
Schedule & IR & FMR & RR \\
\hline
0         & 43.5 & 92.1 & 78.8 \\
5         & 44.1 & 93.1 & 80.5 \\
5 W 10 C  & 44.7 & 93.2 & 82.4 \\
10        & 43.9 & 93.0 & 81.8 \\
10 W 20 C & 45.4 & 93.7 & 83.3 \\
20        & 44.8 & 93.5 & 82.1 \\
\hline\hline
\end{tabular*}
\end{minipage}
\hfill
\begin{minipage}{0.31\textwidth}
\centering
\textbf{C. Comparison with same backbone}
\begin{tabular*}{\linewidth}{@{\extracolsep{\fill}}l|c|c|c}
\hline\hline
Method & IR (\%) & FMR (\%) & RR (\%) \\
\hline
2D3D-MATR & 32.4 & 90.8 & 56.4 \\
CA-I2P    & 35.5 & 93.6 & 63.3 \\
Flow-I2P  & 40.1 & 93.3 & 68.4 \\
GLASS     & 39.2 & 93.3 & 75.3 \\
\hline\hline
\end{tabular*}
\end{minipage}

\label{tab:VII}
\vspace{5pt}
\end{table*}

\subsection{Ablation Studies}
Table~\ref{tab:ablation} presents ablation results of key components in our model. Introducing the LGE module (M2) yields consistent improvements across all metrics, confirming the effectiveness of normal-guided structural enhancement. Adding the GDC module (M3) further boosts PIR and RR, showing the role of structural consistency in reducing mismatches. GAT and DINO (M4–M6) offer moderate gains individually, while their combination (M6) leads to a notable increase in IR and FMR, indicating the benefit of enriched feature representation and attention-based aggregation. The full model (M8), integrating all modules, achieves the best results across all four metrics, validating the complementarity and cumulative effect of each component.
Larger improvements from the ablations reveal that each component targets a different failure mode in cross-modal matching. Specifically, DINO primarily improves semantic discriminability, which enhances coarse correspondence quality (higher PIR/IR) but yields only limited gains in mismatch filtering. In contrast, LGE yields more consistent boosts across PIR/IR/RR by injecting normal-guided geometric cues, aligning local structures, and thus stabilizing pose estimation. GDC contributes most noticeably to FMR and RR, indicating that global structural consistency is crucial for suppressing outliers and reducing mismatches, especially in textureless or repetitive regions. GAT provides moderate gains when used alone, but becomes more effective when combined with DINO, suggesting that attention-based relational aggregation benefits from richer descriptors and leads to better feature representation. The best-performing full model demonstrates that these modules are complementary rather than redundant—local structural enhancement (LGE), global consistency regularization (GDC), relational reasoning (GAT), and semantic enrichment (DINO) jointly produce cumulative improvements.

We highlight the robustness of image normal supervision under noisy depth predictions and conduct controlled ablation experiments to evaluate this issue by injecting two types of realistic perturbations into Depth Anything v2 outputs: (i) Gaussian noise with standard deviations of $\sigma=0.005, 0.01,$ and $0.015$ m, and (ii) random masking of 10\%, 20\%, 30\%, and 40\% of depth pixels to simulate occlusions or low-texture regions. Quantitative results in Table~\ref{tab:VI} A (Gaussian noise) and Table~\ref{tab:VI} B (random masking) show a gradual degradation of all metrics with increasing noise levels: under Gaussian perturbation, IR decreases from 45.4 to 27.9, FMR remains relatively stable (93.7 to 88.5), and RR drops from 83.3 to 51.5 as $\sigma$ rises to 0.015 m; for random masking, IR, FMR, and RR decrease from 45.4 to 33.7, 93.7 to 87.3, and 83.3 to 55.3 respectively with 40\% masking. These results demonstrate our framework is robust to light-to-moderate depth/normal supervision noise (especially FMR), while severe corruption mainly impairs IR and RR (fine-grained keypoint matching and final registration accuracy), clarifying both the robustness and limitations of our method under noisy supervision.

We further conduct an ablation study to explicitly evaluate the impact of incorporating depth cues from Depth Anything v2. The quantitative results are summarized in TABLE~\ref{tab:VI} C. When Depth Anything v2 is included, our method achieves an IR of 46.0, FMR of 94.2, and RR of 84.5. Without depth supervision, the corresponding results are 45.4 (IR), 93.7 (FMR), and 83.3 (RR). Although the performance gap is moderate, consistent improvements are observed across all evaluation metrics. We attribute these gains to the complementary geometric priors provided by Depth Anything v2, which enhance feature discriminability and correspondence reliability, leading to improved inlier ratio and registration robustness. We emphasize that the primary benefit of introducing the Image Normal Prediction Head via distillation is to reduce computational complexity while preserving most of the geometric advantages provided by depth information.

\begin{table*}[!t]
\vspace{-10pt}
\centering
\caption{Results for various values of hyper-parameters on 7-Scenes}
\renewcommand{\arraystretch}{1}

\begin{minipage}{0.32\textwidth}
\centering
\textbf{A. Results for various values of $\lambda_1$}

\begin{adjustbox}{width=\linewidth}
\begin{tabular}{c|c|c|c}
\hline\hline
$\lambda_1$ & IR $\uparrow$ & FMR $\uparrow$ & RR $\uparrow$ \\
\hline
3   & 52.4 & 93.1 & 82.4 \\
2   & 54.1 & 93.7 & 83.1 \\
1   & 54.9 & 93.6 & 84.2 \\
0.5 & 54.8 & 93.2 & 83.4 \\
\hline\hline
\end{tabular}
\end{adjustbox}
\end{minipage}
\hfill
\begin{minipage}{0.32\textwidth}
\centering
\textbf{B. Results for various values of $\lambda_2$}

\begin{adjustbox}{width=\linewidth}
\begin{tabular}{c|c|c|c}
\hline\hline
$\lambda_2$ & IR $\uparrow$ & FMR $\uparrow$ & RR $\uparrow$ \\
\hline
3   & 52.8 & 93.5 & 83.4 \\
2   & 54.3 & 93.2 & 82.1 \\
1   & 54.9 & 93.6 & 84.2 \\
0.5 & 54.1 & 93.4 & 83.8 \\
\hline\hline
\end{tabular}
\end{adjustbox}
\end{minipage}
\hfill
\begin{minipage}{0.32\textwidth}
\centering
\textbf{C. Results for various values of $\lambda_3$}

\begin{adjustbox}{width=\linewidth}
\begin{tabular}{c|c|c|c}
\hline\hline
$\lambda_3$ & IR $\uparrow$ & FMR $\uparrow$ & RR $\uparrow$ \\
\hline
0.7 & 53.8 & 93.5 & 83.5 \\
0.5 & 54.9 & 93.6 & 84.2 \\
0.3 & 54.9 & 93.1 & 84.1 \\
0.1 & 52.5 & 93.0 & 82.0 \\
\hline\hline
\end{tabular}
\end{adjustbox}
\end{minipage}

\label{tab:VIII}
\end{table*}

\begin{table*}[b]
\centering
\caption{Efficiency analysis and ablation study}
\vspace{5pt}
\renewcommand{\arraystretch}{1.1}
\setlength{\tabcolsep}{3pt} 

\begin{minipage}[t]{0.46\textwidth}
\centering
\textbf{A. Efficiency comparison}
\vspace{4pt}
\resizebox{\linewidth}{!}{
\begin{tabular}{c|c|c|c|c}
\hline\hline
Model & 2D3D-MATR & Flow-I2P & Ours & Ours w/o DINOv2 \\
\hline
FPS $\uparrow$ & 7.852 & 6.061 & 1.382 & 5.884 \\
GFLOPs $\downarrow$ & 388.2 & 525.1 & 889.1 & 540.2 \\
Memory (GB) $\downarrow$ & 5.637 & 2.844 & 12.1 & 5.734 \\
Param (M) $\downarrow$ & 28.2 & 28.7 & 322.9 & 29.1 \\
\hline\hline
\end{tabular}
}
\label{tab:efficiency}
\end{minipage}
\hfill
\begin{minipage}[t]{0.36\textwidth}
\centering
\textbf{B. Time comparison}
\vspace{4pt}
\resizebox{\linewidth}{!}{
\begin{tabular}{lcc}
\hline\hline
Module & Matching Time(s) & Total Time(s)\\
\hline
Baseline & 0.038 & 0.694 \\
w DAv2   & 0.080 & 1.837 \\
w/o DAv2 & 0.041 & 0.742 \\
\hline\hline
\end{tabular}
}
\label{tab:time}
\end{minipage}
\hfill
\begin{minipage}[t]{0.14\textwidth}
\centering
\textbf{C. Visualization of ablation on $k$}
\vspace{4pt}
\includegraphics[width=\linewidth]{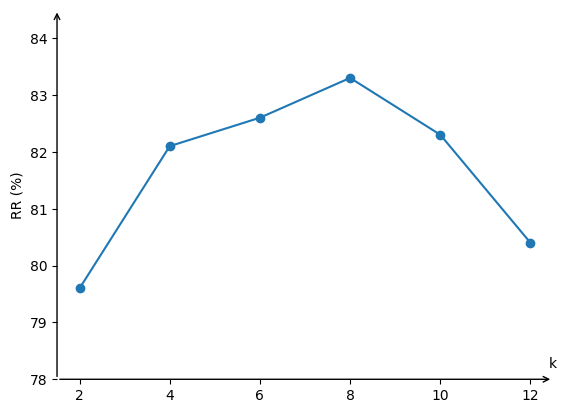} 
\label{tab:IX}
\end{minipage}
\vspace{5pt}
\end{table*}

The neighborhood size $k$ is a critical hyper-parameter for point cloud geometric reasoning. While a fixed setting of $k=8$ demonstrates effectiveness and stability on dense indoor benchmarks (verified via ablation studies), it may lack adaptability in real-world scenarios with significant density variations. To address this limitation, we introduce a simple yet effective density-aware adaptive neighborhood strategy. Specifically, we first compute the local average neighborhood scale $\rho_i$ for each point $x_i$ using an initial value $k_0=8$: $\rho_i = \frac{1}{k_0} \sum_{j=1}^{k_0} \| x_i - x_{ij} \|$, where $x_{ij}$ denotes the $j$-th nearest neighbor of $x_i$. We then dynamically adjust the neighborhood size based on the global mean scale $\text{mean}(\rho)$: if $\rho_i > \text{mean}(\rho)$ (sparse region), $k$ is increased to 12; otherwise (dense region), $k$ remains 8. Experimental results on real-world data, presented in TABLE~\ref{tab:VII} A, confirm that this adaptive strategy consistently improves registration accuracy compared to the fixed $k=8$ setting. It effectively reduces the sensitivity to non-uniform point cloud distributions, leading to more robust feature aggregation in complex outdoor environments.

As an additional analysis of GDC, we examine the early training stage in TABLE~\ref{tab:VII} B, where feature representations are not yet sufficiently discriminative and the similarity distribution is noisy and poorly calibrated. In this regime, aligning similarity distributions may not reliably separate true correspondences from hard negatives, and thus the marginal benefit of GDC can be limited. To verify this, we report results under different training schedules.
From TABLE~\ref{tab:VII} B, directly applying the full objective from scratch yields relatively unstable gains in the early epochs (e.g., IR improves only marginally from 43.5 at epoch 0 to 44.1 at epoch 5, with RR still at 80.5). In contrast, introducing a warm-up stage before enabling the complete objective leads to more consistent improvements. For example, enabling the complete objective after a 5-epoch warm-up (``5 W 10 C'') improves IR/FMR/RR to 44.7/93.2/82.4, and enabling it after a 10-epoch warm-up (``10 W 20 C'') further improves performance to 45.4/93.7/83.3. These results support our observation that GDC is more effective when the feature space has reached a reasonable level of discriminability; otherwise, the similarity structure is too noisy for distribution alignment to provide strong separation.

We also note that different methods employ different backbones. Therefore, in the ablation study of the original paper, we additionally present the comparison results without incorporating DINO v2 features to ensure fairness. For a more intuitive illustration of performance differences, we summarize the results in Table~\ref{tab:VII} C.
It can be observed that even under the same backbone, our method consistently outperforms the baselines across all metrics. In particular, the registration recall (RR) improves from 56.4\% for 2D3D-MATR to 75.3\% for GLASS, representing a gain of approximately 12\% and 7\% compared to CA-I2P and Flow-I2P, respectively. This demonstrates that our approach effectively preserves cross-modal feature consistency while significantly reducing erroneous matches. The improvement is primarily attributed to the local geometry enhancement module, which provides structural cues, and the cross-modal similarity alignment, which further suppresses unreliable correspondences, thereby substantially boosting the final registration performance.

\begin{figure}[!b]
\vspace{-10pt}
\centering
\includegraphics[width=\columnwidth]{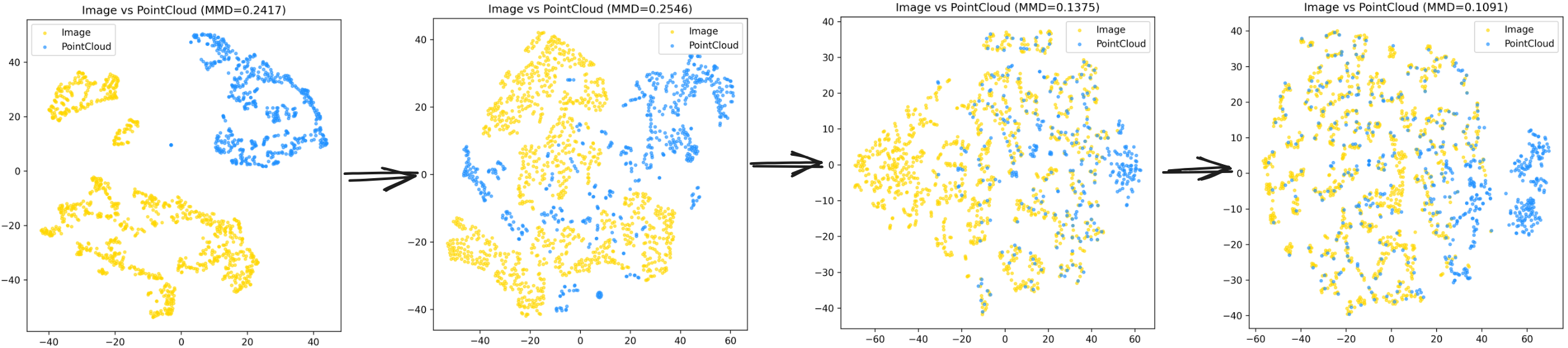}
\caption{Visualization of modality differences. With training, the
modalities and distributions of the point cloud and image become
more aligned.}
\label{tsne}
\end{figure}

The ablation study on the loss coefficients $\lambda_1$, $\lambda_2$, and $\lambda_3$ demonstrates that the proposed method is generally robust to the choice of these hyper-parameters, as shown in Table~\ref{tab:VIII}. Specifically, when varying $\lambda_1$ from 3 to 0.5, the IR changes only from 52.4 to 54.9, the FMR remains within a narrow range of 93.1 to 93.7, and the RR fluctuates slightly between 82.4 and 84.2. A similar trend can be observed for $\lambda_2$, where IR stays between 52.8 and 54.9, FMR remains stable around 93.2 to 93.6, and RR varies from 82.1 to 84.2. For $\lambda_3$, although the performance drops slightly when it is reduced to 0.1, the overall results are still relatively stable across the tested values, with IR ranging from 52.5 to 54.9, FMR from 93.0 to 93.6, and RR from 82.0 to 84.2. Among all settings, the combination around $\lambda_1 = 1$, $\lambda_2 = 1$, and $\lambda_3 = 0.5$ achieves the best overall balance, yielding the highest or near-highest performance on all three metrics. These observations indicate that the proposed method is not overly sensitive to the specific choice of loss coefficients, which confirms the robustness of the optimization objective and suggests that the model can maintain competitive performance under different hyper-parameter configurations.

We conduct a more detailed efficiency analysis. As shown in Table~\ref{tab:IX} A, our method's primary computational overhead originates from the use of the DINOv2 backbone for feature extraction. With DINOv2 included, our full model achieves the best registration performance, but this comes at a clear efficiency cost: the inference speed drops to 1.382 FPS, while the computational cost, memory usage, and parameter count increase to 889.1 GFLOPs, 12.1 GB, and 322.9M, respectively.
By contrast, removing DINOv2 substantially improves efficiency. The frame rate increases from 1.382 FPS to 5.884 FPS, GFLOPs decrease from 889.1 to 540.2, memory usage drops from 12.1 GB to 5.734 GB, and the parameter count is reduced dramatically from 322.9M to 29.1M. Importantly, under this setting, the efficiency of our method is comparable to that of existing approaches such as 2D3D-MATR and Flow-I2P. For example, the model size is nearly the same as 2D3D-MATR and Flow-I2P, and both the memory consumption and runtime are within a competitive range.

\begin{figure*}[t]
\centering
\includegraphics[width=\textwidth]{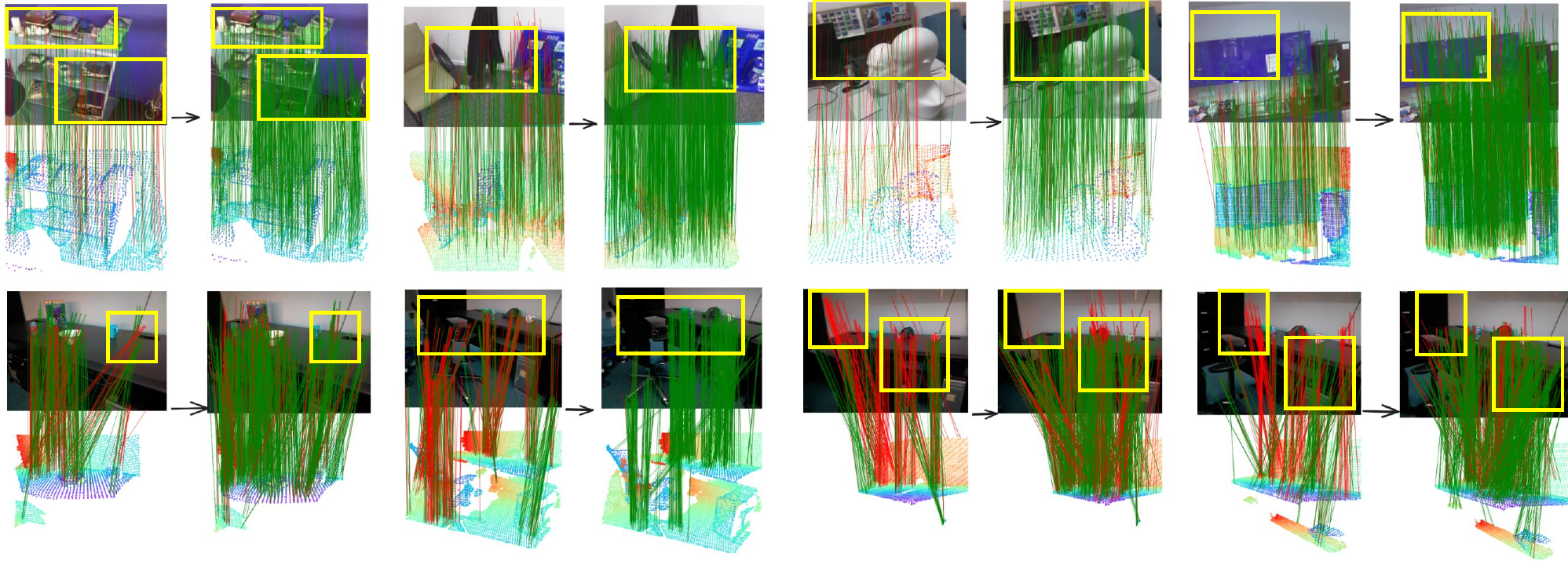}
\caption{Visualization of the image-to-point cloud matching results of GLASS. To rigorously analyze the performance, we set the error threshold to a strict 30px. As seen, our method significantly improves the matching performance in challenging scenarios.}
\vspace{-10pt}
\label{vis}
\end{figure*}

\begin{figure}[!b]
\vspace{-10pt}
\centering
\includegraphics[width=\columnwidth]{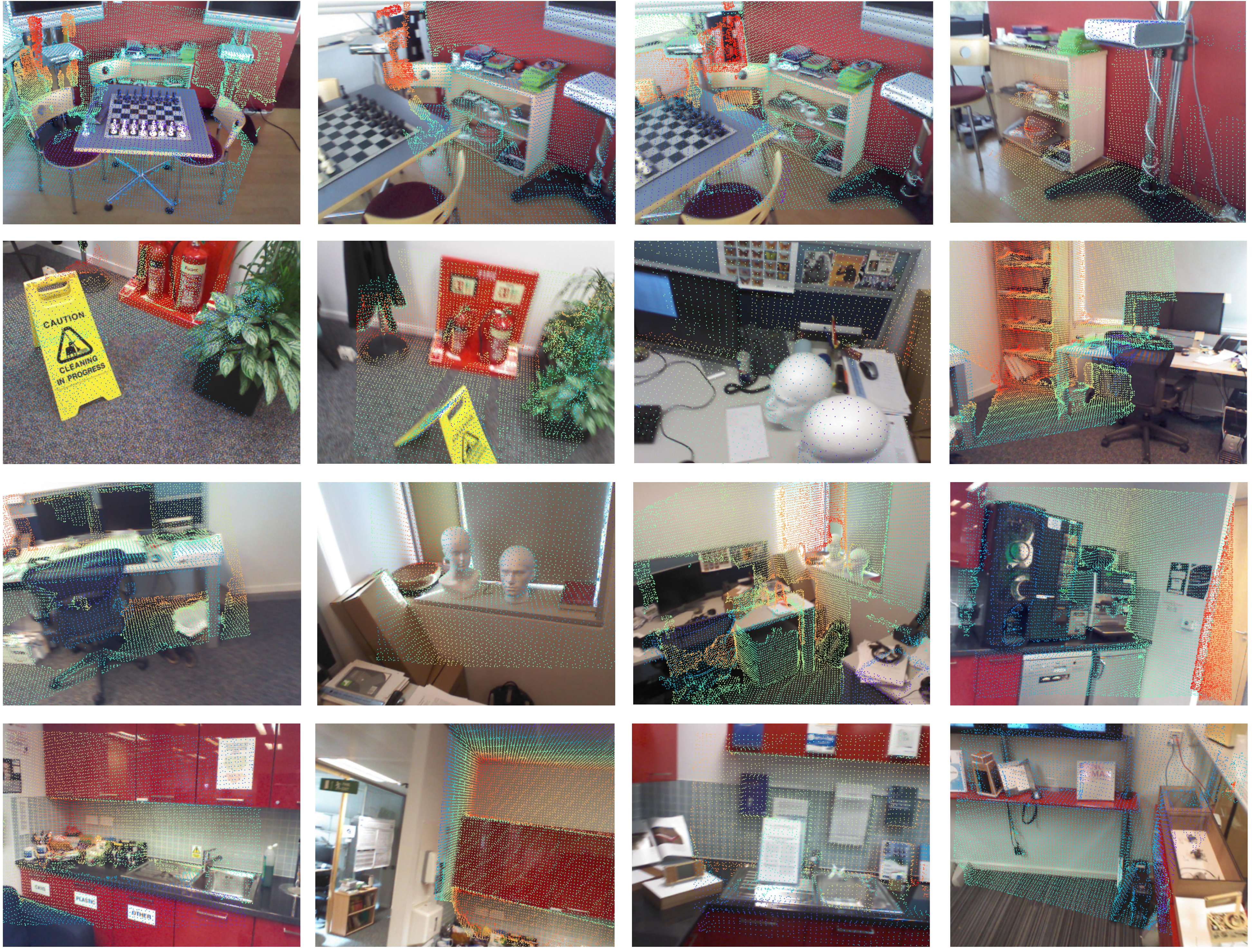}
\caption{The visualization of the point cloud projection onto the image shows that our method achieves an accurate rigid transformation, without causing significant misalignment.}
\label{vis2}
\end{figure}

This analysis indicates that while DINOv2 provides high-quality, discriminative features that improve matching and registration performance, it comes at a significant computational cost. Therefore, there exists a clear trade-off between accuracy and efficiency: retaining DINOv2 maximizes registration precision but reduces inference speed, whereas removing it greatly improves real-time applicability at the expense of some accuracy. We will include this discussion in the revised manuscript to provide a more complete view of computational complexity, memory consumption, and real-time performance considerations.

To assess the computational impact of Depth Anything v2 (DAv2), we compare matching and total inference times with and without DAv2 (Table~\ref{tab:IX} B). DAv2 increases the matching time from 0.041s to 0.080s and raises the total inference time from 0.742s to 1.837s, indicating a significant runtime overhead. To maintain efficiency, we disable DAv2 during inference. Compared to our baseline, GLASS incurs a slight increase in computational complexity. However, considering the significant improvement in performance, this trade-off is justified.

The value of \( k \) determines the range of surrounding information aggregation. A too-small range may result in the loss of surrounding information, while a too-large range can lead to the incorrect merging of different classes. Through our ablation study on RGB-D v2 (referenced in Table~\ref{tab:IX} C), we found that the best performance is achieved when \( k = 8 \), balancing the range of aggregation effectively without losing important details or causing incorrect class fusion.

\subsection{Visualization}
Fig.~\ref{tsne} visualizes the distributional differences between image and point cloud features using t-SNE at different training stages. At the beginning of training, the two modalities are clearly separated, reflecting a large domain gap between image and point cloud representations. As training progresses, their feature distributions gradually become more mixed and overlapping, indicating that our model effectively reduces the modality discrepancy and achieves better cross-modal alignment. The decreasing MMD values shown above each figure further confirm that the domain gap is progressively minimized.

Fig.~\ref{vis} visualizes the matching results of GLASS. A match is considered correct if the projected point cloud keypoint lies within 30 pixels of its image counterpart—used to distinguish correct (green) from incorrect (red) matches. After structured modeling, our method significantly improves accuracy by producing more correct correspondences and fewer false matches, as reflected by the increased green and reduced red lines. The reduction in redundant matches also leads to more robust and reliable registration.

In Fig.~\ref{vis2}, we project the point clouds onto the images using the estimated poses, showing no large-scale misalignment across scenes, indicating satisfactory registration.

\begin{table}[b]
\caption{Comparison of RR Performance on Pumpkin and Stairs Scenes}
\centering
\resizebox{0.65\columnwidth}{!}{
\begin{tabular}{c|c|c}
\hline\hline
\textbf{Scenes} & \textbf{Pumpkin} & \textbf{Stairs} \\
\hline
Previous & 83.8 & 29.7 \\
\hline
Modify   & 84.3 & 36.8 \\
\hline\hline
\end{tabular}
}
\label{limit}
\end{table}

\subsection{Limitation and potential improvements}

While our method makes significant strides in advancing image-to-point cloud registration, some long-standing challenges in the field remain. We have successfully addressed several of these, such as improving domain alignment and reducing mismatches, but there is still room for further refinement. In this section, we discuss these ongoing challenges and propose potential future directions to enhance the robustness and accuracy of our approach.

\begin{figure}[t]
\centering
\includegraphics[width=\columnwidth]{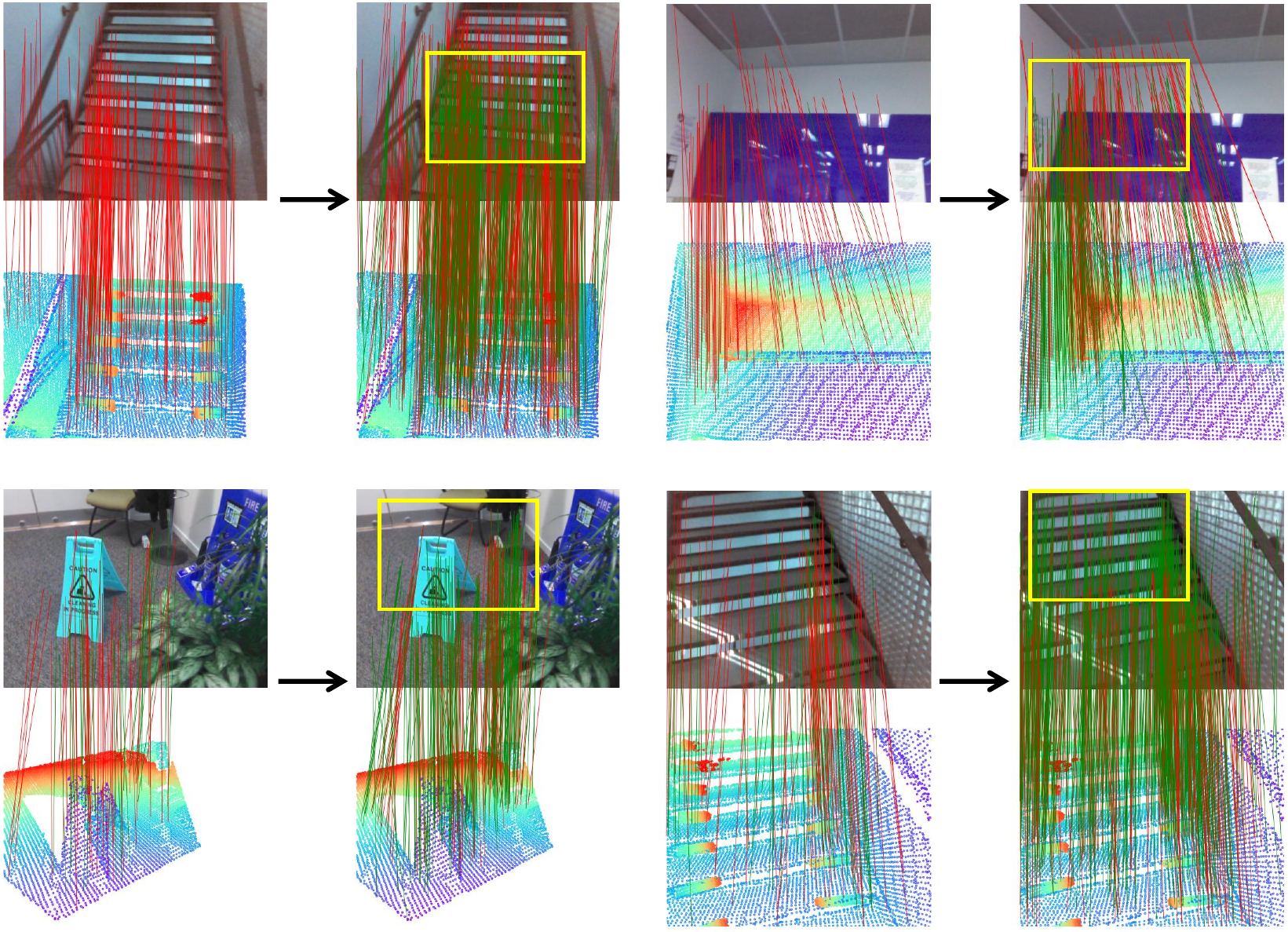}
\caption{Limitation visualization on Dataset. The figure illustrates the comparison from the baseline to our method in challenging regions.    
The comparison from left to right represents the baseline and our method.}
\vspace{-10pt}
\label{limita}
\end{figure}

To better understand the limitations of our method and explore potential avenues for improvement, we visualize several challenging cases. As shown in Fig.~\ref{limita}, while our approach outperforms previous baselines in these scenarios, some incorrect correspondences persist. We observe that in regions with rapid depth changes—such as stairway gaps, ceiling vents, and hanger bases—normals fail to provide reliable information and may even contribute to incorrect correspondences. We suggest potential solutions to address this issue. Given that the ultimate goal of feature matching is to ensure reliable correspondences for the subsequent PnP-RANSAC algorithm, our efforts should focus on improving correspondence accuracy.

Indeed, since our framework leverages Depth Anything v2 to provide geometric cues, the quality of the depth map may influence the behavior of the proposed LGE module. Regarding the impact of depth-map artifacts or erroneous structures on the LGE module, our failure-case visualizations (Fig.~\ref{limita}) show that the main issues typically occur in regions where establishing stable correspondences is inherently difficult, such as low-texture areas, repetitive patterns, occlusion or reflection, thin structures, or large viewpoint changes. These regions are prone to matching ambiguity even without using depth input. While local artifacts from Depth Anything v2 may reduce geometric consistency, the depth cues still provide useful local geometric guidance, which in our method helps increase both the number of matches and the matching accuracy during the correspondence stage. As shown in Fig.~\ref{limita}, although mismatches still exist in regions with dense depth variations (e.g., stair shadows), such errors were already present in the baseline method. In contrast, our approach not only produces more correspondences (denser match connections), but also substantially improves the correctness of matches in regions such as stair gaps and walls. For the final registration, we employ a PnP+RANSAC post-processing step; therefore, even if a portion of correspondences are incorrect, they can be effectively suppressed by the robust estimation process.

As a potential future direction, we plan to integrate an uncertainty-aware local geometry modeling strategy. By explicitly modeling depth uncertainty, unreliable regions (e.g., stair gaps and ceiling vents) can be down-weighted, while local geometric fitting methods (e.g., plane or curvature regularization) can be employed to refine normal estimation. This approach may help stabilize normals in discontinuous regions and further reduce incorrect correspondences. We tested our method on the existing challenging scenarios, Pumpkin and stairs, and found significant improvements in performance (shown in Table~\ref{limit}).

\section{Conclusion}
In this paper, we introduce an innovative geometry-aware local alignment and structure synchronization network for 2D-3D registration.
Our method efficiently extracts surface normal information from 2D images to capture rich structural cues, which enhances the correlation with point cloud features and effectively reduces cross-modal mismatches. Additionally, the use of graph-based feature propagation and structural similarity constraints improves the consistency of matched correspondences, significantly boosting registration accuracy and robustness. Extensive experiments on the RGB-D Scenes v2 and 7-Scenes datasets demonstrate that our GLASS approach achieves state-of-the-art performance in image-to-point cloud registration.

\begin{IEEEbiography}[{\includegraphics[width=1in,height=1.25in,clip,keepaspectratio]{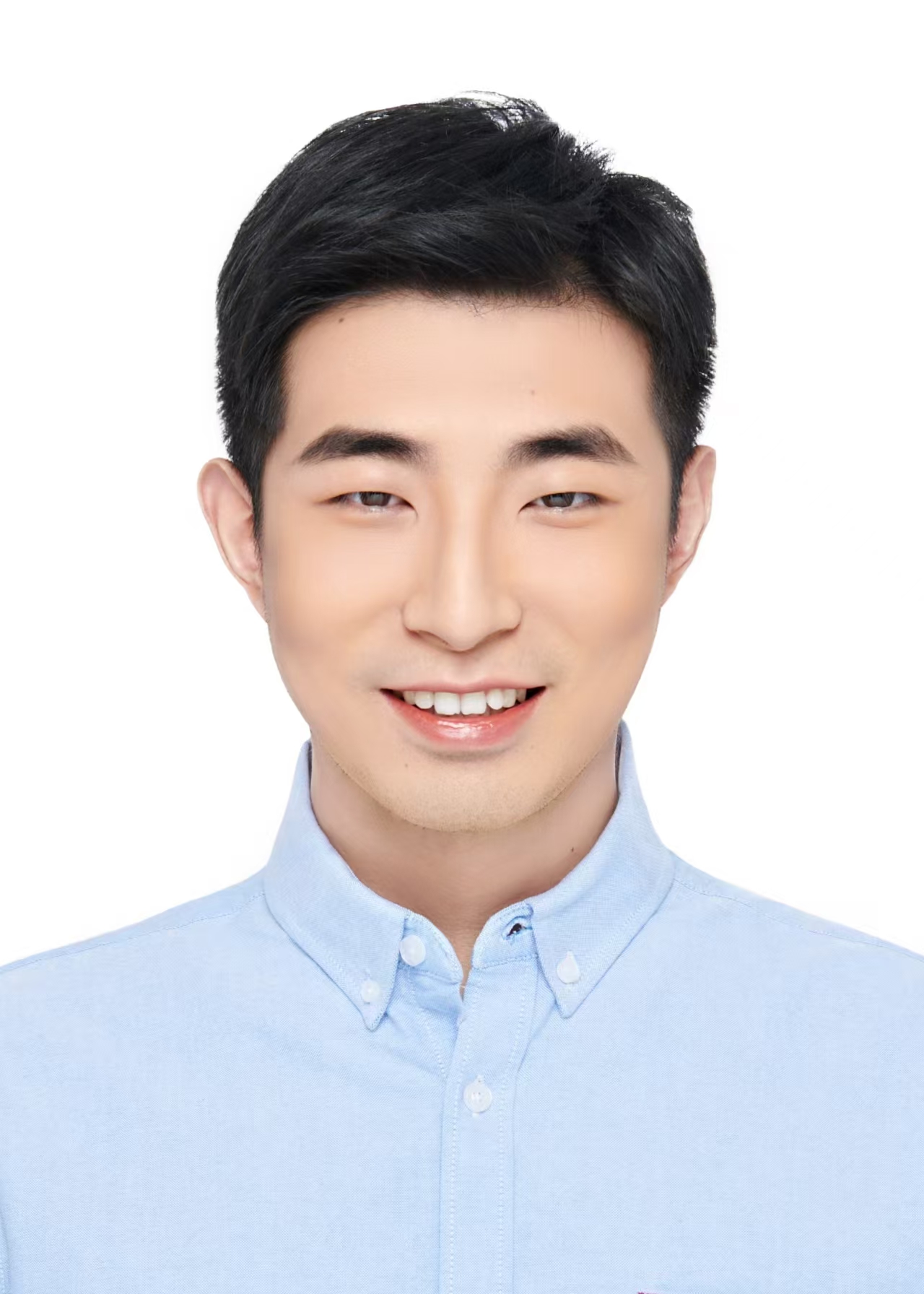}}]{Zhixin Cheng}
 received his bachelor's degree from the School of Electrical and Electronic Engineering, Huazhong University of Science and Technology, Wuhan, China, in 2020. He is currently pursuing a Ph.D. degree at the School of Information Science and Technology, University of Science and Technology of China, Hefei, China. His main research interests include computer vision and machine learning, with a focus on 3D scene registration task and multi-modal learning.
\end{IEEEbiography}

\begin{IEEEbiography}[{\includegraphics[width=1in,height=1.25in,clip,keepaspectratio]{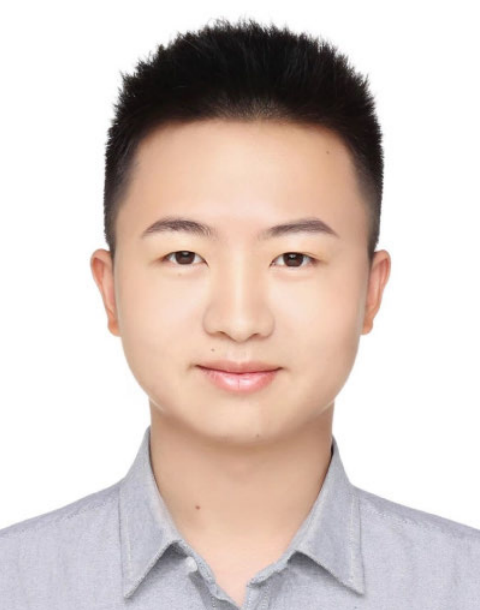}}]{Jiacheng Deng}
  received the bachelor's degree in Information Security from the University of Science and Technology of China in 2023. He received the Ph.D. degree in Control Science and Engineering from the University of Science and Technology of China. He is currently working at Meituan. His research interests include computer vision and deep learning, especially image-to-point cloud registration and pose estimation.
\end{IEEEbiography}

\begin{IEEEbiography}[{\includegraphics[width=1in,height=1.25in,clip,keepaspectratio]{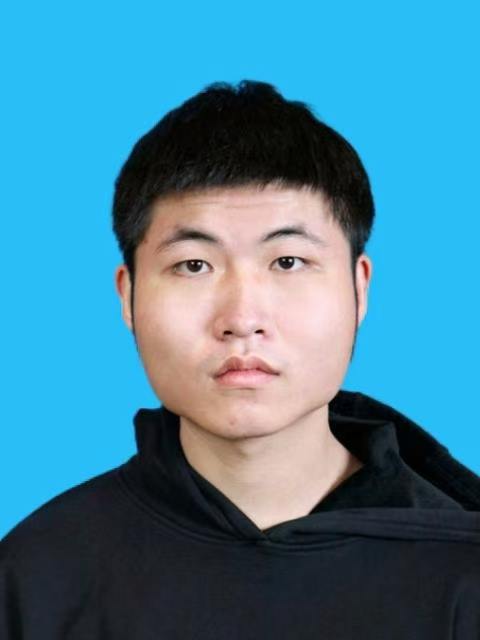}}]{Xinjun Li}
  received the bachelor’s degree in Mathematics and Applied Mathematics from the University of Science and Technology of China, Hefei, China, in 2020. He is currently pursuing a Ph.D. degree in Control Science and Engineering at the same university. His research interests include computer vision and machine learning, with a focus on 3D object detection, shape correspondence, and point cloud segmentation.
\end{IEEEbiography}

\begin{IEEEbiography}[{\includegraphics[width=1in,height=1.25in,clip,keepaspectratio]{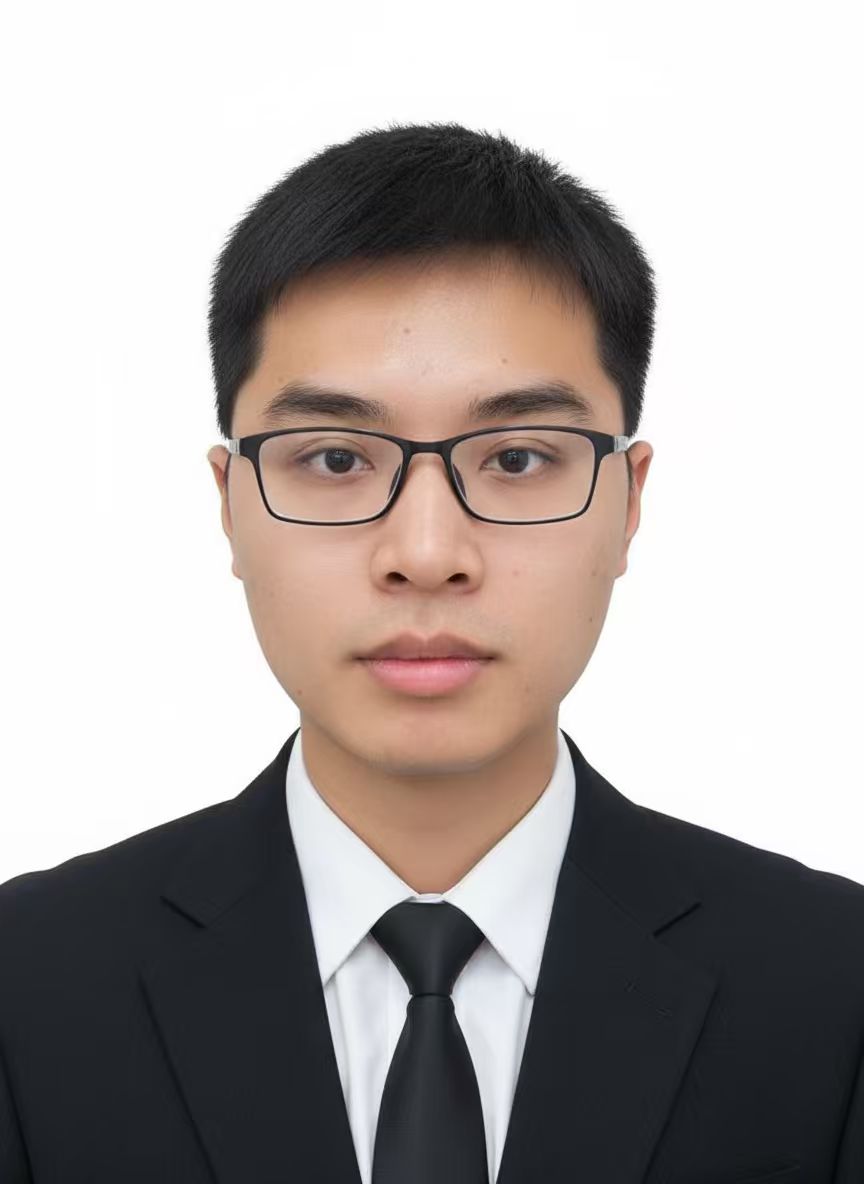}}]{Bohao Liao} received the B.E. degree from the
University of Science and Technology of China,
Hefei, China, in 2022, where he is currently pursuing
the Ph.D. degree.
His current research interests include machine
learning and computer vision.
\end{IEEEbiography}

\begin{IEEEbiography}[{\includegraphics[width=1in,height=1.25in,clip,keepaspectratio]{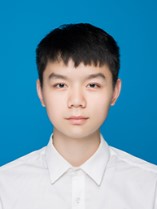}}]{Li Liu}
   received the bachelor’s degree in Computer Science from Hefei University of Technology, China, in 2022. He is currently pursuing a master’s degree in Electronic and Information Engineering at the University of Science and Technology of China. His research interests include computer vision and machine learning, with a focus on depth estimation and stereo matching.
\end{IEEEbiography}

\begin{IEEEbiography}[{\includegraphics[width=1in,height=1.25in,clip,keepaspectratio]{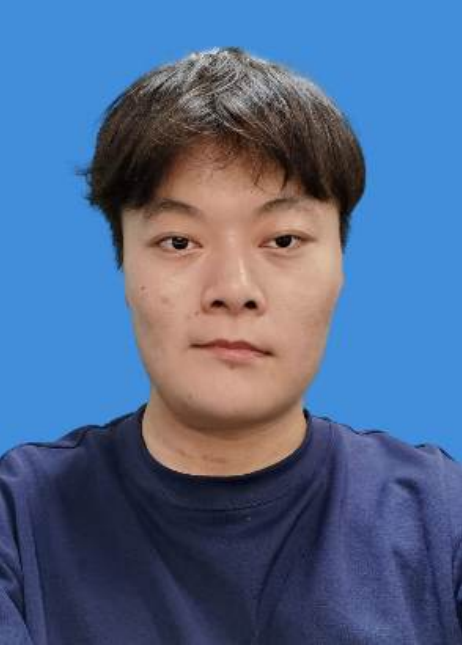}}]{Xiaotian Yin}
    is currently pursuing a Ph.D. degree at the University of Science and Technology of China, Hefei, China. His research interests include computer vision and machine learning, with a focus on few-shot learning and multi-modal learning.
\end{IEEEbiography}

\begin{IEEEbiography}[{\includegraphics[width=1in,height=1.25in,clip,keepaspectratio]{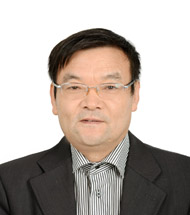}}]{Baoqun Yin}
   received his bachelor’s degree in Mathematics from Sichuan University, Chengdu, China, in 1985, and his master’s degree in Applied Mathematics from the University of Science and Technology of China (USTC), Hefei, China, in 1993. He earned his Ph.D. degree in Pattern Recognition and Intelligent Systems from the Department of Automation, USTC, in 1998. He is currently a Professor in the Department of Automation at the University of Science and Technology of China. His research interests include stochastic systems, system optimization, and information networks, focusing on Markov decision processes, network optimization, and smart energy management.
\end{IEEEbiography}

\begin{IEEEbiography}[{\includegraphics[width=1in,height=1.25in,clip,keepaspectratio]{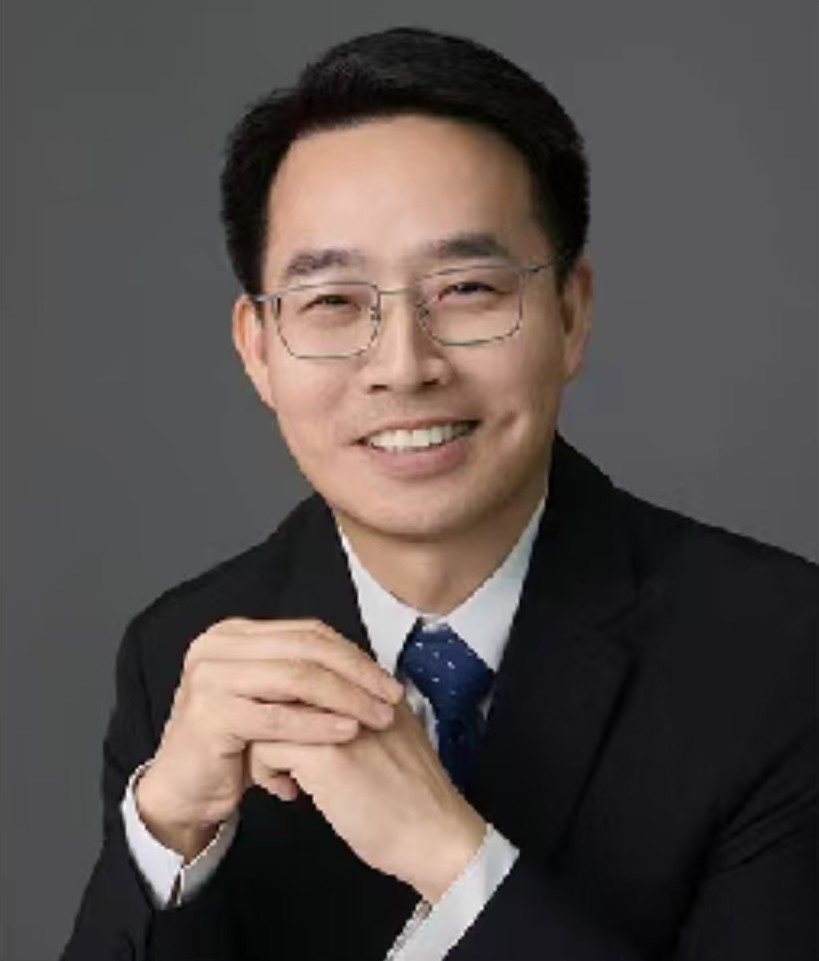}}]{Tianzhu Zhang}
  (M’11) received the bachelor’s degree in Communications and Information Technology from the Beijing Institute of Technology, Beijing, China, in 2006, and the Ph.D. degree in Pattern Recognition and Intelligent Systems from the Institute of Automation, Chinese Academy of Sciences, Beijing, China, in 2011. He is currently a Professor at the Department of Automation, School of Information Science, University of Science and Technology of China. His current research interests include computer vision and multimedia, with a focus on action recognition, object classification, object tracking, and social event analysis.
\end{IEEEbiography}

\vfill

\end{document}